\documentclass{article}

\usepackage[utf8]{inputenc}

\usepackage[hidelinks]{hyperref}

\usepackage[accepted]{icml2020}

\usepackage{dsfont}
\usepackage{url}
\usepackage{booktabs}       
\usepackage{amsfonts}       
\usepackage{amsmath}
\usepackage{amssymb}
\usepackage{amsthm}
\usepackage{cancel}
\usepackage{hyphenat}
\usepackage[stretch=15, shrink=15, factor=1200]{microtype}

\usepackage{nicefrac}       
\usepackage{float}
\usepackage{subcaption}
\usepackage{enumitem}
\usepackage{listings}
\usepackage{xcolor}
\usepackage{tikz}
\usepackage{graphicx}

\usetikzlibrary{calc}

\makeatletter
\newcommand{\shorteq}{%
  \settowidth{\@tempdima}{-}
  \resizebox{\@tempdima}{\height}{=}%
}
\makeatother

\makeatletter
\newcommand{\shortminus}{%
  \settowidth{\@tempdima}{-}
  \resizebox{\@tempdima}{\height}{=}%
}
\makeatother

\definecolor{dark-red}{rgb}{0.8, 0.1, 0.1}

\newcommand\undermat[2]{%
  \makebox[0pt][l]{$\smash{\underbrace{\phantom{%
    \begin{matrix}#2\end{matrix}}}_{\text{$#1$}}}$}#2}

\definecolor{dark-red}{rgb}{0.8, 0.1, 0.1}

\usepackage{amsmath,amsfonts,bm}

\def\eqref#1{equation~\ref{#1}}
\def\Eqref#1{Equation~\ref{#1}}

\def\1{\bm{1}}

\def\va{{\bm{a}}}
\def\vb{{\bm{b}}}
\def\vc{{\bm{c}}}
\def\vd{{\bm{d}}}

\def\vh{{\bm{h}}}

\def\vm{{\bm{m}}}

\def\vs{{\bm{s}}}

\def\vv{{\bm{v}}}
\def\vw{{\bm{w}}}
\def\vx{{\bm{x}}}
\def\vy{{\bm{y}}}

\DeclareMathAlphabet{\mathsfit}{\encodingdefault}{\sfdefault}{m}{sl}
\SetMathAlphabet{\mathsfit}{bold}{\encodingdefault}{\sfdefault}{bx}{n}

\newcommand{\E}{\mathbb{E}}

\newcommand{\R}{\mathbb{R}}

\newtheorem{theorem}{Theorem}

\theoremstyle{definition}

\DeclareMathOperator{\cov}{cov}
\DeclareMathOperator{\diag}{diag}

\newcommand{\norm}[1]{\left\lVert#1\right\rVert}

\definecolor{anti-flashwhite}{rgb}{0.95, 0.95, 0.96}
\usepackage[disable]{todonotes}

\newcommand{\indicator}{1 
}
\newcommand{\abLRP}{LRP$_{\alpha\beta}$}
\newcommand{\alphaBetaLRP}[2]{LRP$_{\alpha{#1}\beta{#2}}$}
\newcommand{\aIbOLRP}{\alphaBetaLRP{1}{0}}

\newcommand{\zLRP}{LRP$_z$}

\newcommand{\idx}[1]{{[#1]}}
\newcommand{\scos}{s_{\cos}}
\newcommand{\LRPcmp}{LRP$_{CMP}$}
\newcommand{\vgamma}{\bm{\gamma}}

\begin{document}

\twocolumn[
\icmltitle{%
    When Explanations Lie: Why Many Modified BP Attributions Fail
}



\icmlsetsymbol{equal}{*}

\begin{icmlauthorlist}
\icmlauthor{Leon Sixt}{fu}
\icmlauthor{Maximilian Granz}{fu}
\icmlauthor{Tim Landgraf}{fu}
\end{icmlauthorlist}

\icmlaffiliation{fu}{Dahlem Center of  Machine Learning and Robotics, Freie Universität Berlin, Germany}

\icmlcorrespondingauthor{Leon Sixt}{leon.sixt@fu-berlin.de}

\icmlkeywords{Attribution, Explainable AI, Interpretable Machine Learning, Modified Backpropagation, Sanity Checks, LRP}

\vskip 0.3in
]



\printAffiliationsAndNotice{} 

\begin{abstract}
    Attribution methods aim to explain a neural network's prediction by highlighting the most relevant image areas. A popular approach is to backpropagate (BP) a custom relevance score using modified rules, rather than the gradient. We analyze an extensive set of modified BP methods: Deep Taylor Decomposition, Layer-wise Relevance Propagation (LRP), Excitation BP, PatternAttribution, DeepLIFT, Deconv, RectGrad, and Guided BP. We find empirically that the explanations of all mentioned methods, except for DeepLIFT, are independent of the parameters of later layers. We provide theoretical insights for this surprising behavior and also analyze why DeepLIFT does not suffer from this limitation. Empirically, we measure how information of later layers is ignored by using our new metric, cosine similarity convergence (CSC). The paper provides a framework to assess the faithfulness of new and existing modified BP methods theoretically and empirically.
    \footnote[2]{
    For code see: \href{https://github.com/berleon/when-explanations-lie}{github.com/berleon/when-explanations-lie}

}
\end{abstract}

\section{Introduction}

Explainable AI (XAI) aims to improve the interpretability of machine learning models. For deep convolutional networks, attribution methods visualize the areas relevant for the prediction with so-called saliency maps. Various attribution methods have been proposed, but do they reflect the model behavior correctly?

\begin{figure}[t]
    \centering

     \begin{subfigure}[t]{\linewidth}
      \centering
        \includegraphics[width=0.88\linewidth]{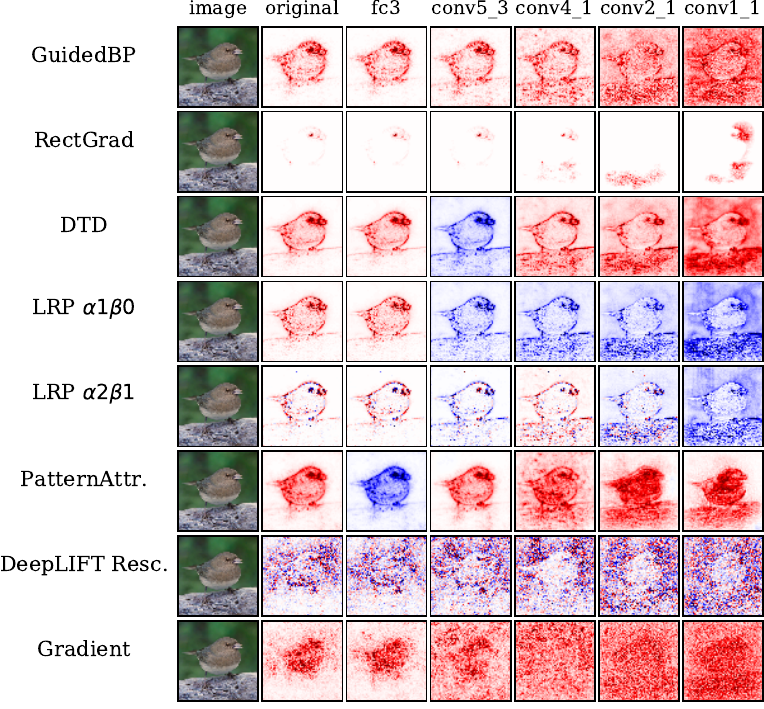}
        \caption{Sanity Checks (VGG-16)}
        \label{fig:figure1_sanity_checks}
    \end{subfigure}

    \newcommand{\dogcatheight}{1.6cm}
    \newcommand{\dogcatwidth}{0.22\linewidth}
    \begin{subfigure}[t]{\dogcatwidth}
     \centering
        \includegraphics[height=\dogcatheight]{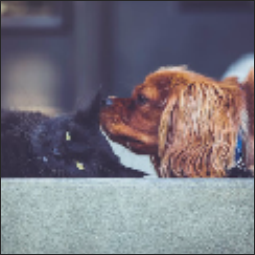}
        \caption{Image}
        \label{fig:figure1_cat_dot}
    \end{subfigure}
    \begin{subfigure}[t]{\dogcatwidth}
     \centering
        \includegraphics[height=\dogcatheight]{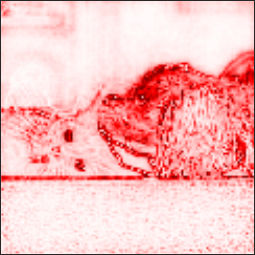}
        \caption{Expl. Cat}
    \end{subfigure}
    \begin{subfigure}[t]{\dogcatwidth}
     \centering
        \includegraphics[height=\dogcatheight]{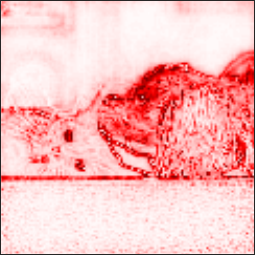}
        \caption{Expl. Dog}
    \end{subfigure}
    \begin{subfigure}[t]{0.31\linewidth}
     \centering
        \includegraphics[height=\dogcatheight]{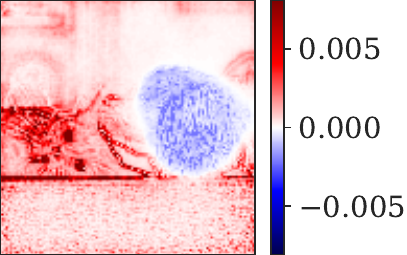}
        \caption{Diff. (c) - (d)}
        \label{fig:figure1_diff_cat_dog}
    \end{subfigure}
    \caption{\textbf{(a)} Sanity Checks: Saliency maps should change if network parameters are randomized. Parameters are randomized from the last to the first layer. \emph{Red} denotes positive and \emph{blue} negative relevance. \textbf{(b-e)} Class insensitivity of \aIbOLRP{}
    on VGG-16. Explanation for \textbf{(c)} \emph{Persian cat (283)} and \textbf{(d)} \emph{King Charles Spaniel (156)}. \textbf{(e)} Difference (c) - (d), both normalized to $[0, 1]$. L1-norm of (e) = 0.000371.
   }
    \label{fig:figure1}
\end{figure}

\cite{adebayo2018sanity} proposed a sanity check: if the parameters of the model are randomized and therefore the network output changes, do the saliency maps change too? Surprisingly, the saliency maps of {GuidedBP} \citep{springberg2014gbp} stay identical, when the last layer (fc3) is randomized (see Figure \ref{fig:figure1_sanity_checks}). A method ignoring the last layer can \emph{not} explain the network's prediction faithfully.

In addition to \cite{adebayo2018sanity}, which only reported GuidedBP to fail,
we found several modified backpropagation (BP) methods fail too: {Layer-wise Relevance Propagation (\textsc{LRP})}, Deep Taylor Decomposition \textsc{(DTD)}, {PatternAttribution}, {Excitation BP},  {Deconv}, {GuidedBP}, and {RectGrad} \citep{bach2015lrp,montavon2017dtd,kindermans2018learning,zhang2018excitationbp,zeiler2014deconvnet,springberg2014gbp,kim2019rectgrad}.
The only tested modified BP method passing is DeepLIFT \citep{shrikumar2017deeplift}.

Modified BP methods estimate relevant areas by backpropagating a custom relevance score instead of the gradient. For example, DTD only backpropagates positive relevance scores. Modified BP methods are popular with practitioners \citep{yang2018therapy,sturm2016interpretable,eitel2019ms}.
For example, \cite{schiller2019whale} uses saliency maps to improve the classification of whale sounds
or \cite{bohle2019lrp} use \aIbOLRP{} to localize evidence for Alzheimer's disease in brain MRIs.

Deep neural networks are composed of linear layers (dense, conv.) and
non-linear activations. For each linear layer, the weight vector reflects the
importance of each input directly.
\citep{bach2015lrp,kindermans2018learning,montavon2017dtd} argue that
aggregating explanations of each
linear model can explain a a deep neural network. Why do these methods then fail the sanity check?

Theoretically, we show that the $z^+$-rule -- used by DTD, \aIbOLRP, and Excitation BP -- yields a multiplication chain of non-negative matrices. Each matrix corresponds to a layer.
The saliency map is a function of this matrix chain.
We show that such a  non-negative matrix chain converges to a rank-1 matrix. If $C\! \in\! \R^{n\times m}$ is a rank-1 matrix, then it can be written as an outer product $C\! =\!\vc \vgamma^T,\ \vc\!\in\!\R^n$, $\vgamma\! \in\! \R^m$. Multiplying $C$ with any vector $\vv$ yields always the same the direction: $C \vv=\vc \vgamma^T \vv=\!\lambda \vc, \lambda\! \in\! \R$.
The scaling is irrelevant as saliency maps are normalized.
If sufficiently converged, the backpropagated vector can merely switch the sign of the saliency map.
For example, in Figure \ref{fig:figure1_sanity_checks}, the sign of the PatternAttribution saliency map
switches due to  the randomization of fc3. Figure \ref{fig:figure1_cat_dot}-\ref{fig:figure1_diff_cat_dog} show how
the saliency maps of \aIbOLRP{} become class-insensitive.

Empirically, we quantify the convergence to a rank-1 matrix using our novel cosine similarity convergence (CSC) metric. CSC allows to retrace, layer by layer, how modified BP methods lose information about previous layers. Using CSC, we observe that all analyzed modified BP methods, except for DeepLIFT, converge towards a rank-1 matrix on VGG-16 and ResNet-50. For sufficiently large values of $\alpha$ and $\beta$, \abLRP\ does not converge but also produces rather noisy saliency maps.

The paper focuses on modified BP methods, as other attribution methods do not suffer from the converges problem. They either rely on the gradient directly \cite{smoothgrad,sundararajan2017axiomatic}, which does not converge or consider the model as a black-box \cite{ribeiro2016lime,lundberg2017shap}.

Our findings show that many modified BP methods are prone to class-insensitive explanations and provide saliency maps that rather highlight low-level features. Negative relevance scores are crucial to avoid the convergence to a rank-1 matrix --- a possible future research direction.

\section{Theoretical Analysis}

\paragraph{Notation}
For our theoretical analysis, we consider feed-forward neural networks with a ReLU activation function  $[\vx]^+ = \max(0, \vx)$. The neural network $f(\vx)$
contains $n$ layers, each with weight matrices $W_l$. The output of the $l$-th layer is denoted by  $\vh_l$.
We use $\idx{ij}$ to index the $i, j$ element in $W_l$ as in $W_{l_\idx{ij}}$.
To simplify notation, we absorb the bias terms into the weight matrix, and we omit the final softmax layer. We refer to the input with $\vh_0 = \vx$ and to the output with $\vh_n = f(\vx)$. The output of the $l$-th layer is given by:
\begin{equation}
    \vh_l = [W_l \vh_{l-1}]^+
\end{equation}
All the results apply to convolutional neural networks as convolution can be expressed as matrix multiplication.

\paragraph{Gradient} The gradient of the $k$-th output of the neural network w.r.t. the input $\vx$ is given by:
\begin{equation}
     \frac{\partial f_k(\vx)}{\partial \vx}  =
     W_1^T  M_1 \frac{\partial f_k(\vx)}{\partial \vh_{1}}
     =\prod_l^n \left( W_l^T M_l \right) \cdot \vv_k,
\end{equation}
where $M_l = \diag(\indicator_{h_l > 0})$ denotes the gradient mask of the ReLU operation. The last equality follows from recursive expansion. The vector $\vv_k$ is a one-hot vector to select the $k$-th output.

The gradient of residual blocks is also a product of matrices. The gradient of $\vh_{l+1} = \vh_l + g(\vh_l)$ is:
\begin{equation}
    \frac{\partial \vh_{l+1}}{\partial \vh_l} =  I + G_{\partial g(\vh_l) / \partial \vh_l},
\end{equation}
where $G_{\partial g(\vh_l) / \partial \vh_l}$ denotes the derivation matrix of the residual block, and $I$ is the identity matrix.
For the gradient, the final saliency map is usually obtained by summing the absolute channel values of the relevance vector $r^\nabla_0(\vx)$ of the input layer.

The following methods modify the gradient definition and to distinguish the rules,
we introduce the notation: $r_l^\nabla(\vx)= \frac{\partial f(\vx)}{\partial \vh_l}$ \todo{R9: not used later}
which denotes the relevance at layer $l$ for an input $\vx$.

\paragraph{Interpretability of Linear Models}
The relevance of the input of a linear model can be calculated directly.
Let $y = \vw^T \vx$ be a linear model with a single output scalar. The relevance of the input $\vx$ to the $i$-th output $\vy_\idx{i}$ is :
\begin{equation}
    r_\vx^{\text{Linear}}(\vx) = \vw \odot \vx.
\end{equation}

\subsection{$z^+$-Rule}
The $z^+$-rule is used by DTD \citep{montavon2017dtd}, Excitation BP \citep{zhang2018excitationbp} and also corresponds to the \alphaBetaLRP{1}{0} \ rule \citep{bach2015lrp}. The $z^+$-rule backpropagates positive relevance values, which are supposed to correspond to the positive evidence for the prediction. Let $w_{ij}$ be an entry in the weight matrix $W_l$:
\newcommand{\rzplus}[1]{r_{#1}^{z^+}}
\begin{equation}
\begin{split}
    \rzplus{l}(\vx) &= Z^+_l \cdot \rzplus{l+1}(\vx) \\
    \text{where} \, Z_{l}^{+^T} &= \left(
                \frac{[w_{ij} \vh_{l_\idx{j}}]^+}
                     {\sum_k [w_{ik} \vh_{l_\idx{k}}]^+}
            \right)_\idx{ij}
\end{split}
\end{equation}
Each entry in the derivation matrix $Z^+_l$ is obtained by measuring the positive contribution of the input neuron $i$ to the output neuron $j$ and normalizing by the total contributions to neuron $j$. The relevance from the previous layer $\rzplus{l+1}$ is then distributed according to $Z^+_l$.
The relevance function $\rzplus{l}: \R^n \mapsto \R^m $
maps input $\vx$ to a relevance vector of layer $l$. For the final layer the relevance is set to the value of the explained logit value, i.e. $\rzplus{n}(\vx) = f_k(\vx) $.
In contrast to the vanilla backpropagation, algorithms using the $z^+$-rule do not apply a mask for the ReLU activation. \todo{L: check how the []+ replaces the relu}

The relevance of multiple layers is computed by applying the $z^+$-rule to each of them. Similar to the gradient, we obtain a product of non-negative matrices: $C_k = \prod^k_l Z^+_l$.

\newcommand{\epsDot}{\epsilon_{\langle \cdot,  \cdot \rangle } }

\begin{theorem}\label{Thm1}
Let $A_1, A_2, A_3\dots$ be a sequence of non-negative matrices. 
We require that every column vector $\va$ of $A_n$ has a norm $|| \va || \ge \epsilon_0$
and that infinite many matrices $A_i$ with $i \in I$ and $|I| = |\mathbb{N}|$ exists for which two column vectors have a dot product of at least $\epsilon_{\langle \cdot,  \cdot \rangle }$, i.e. 
$\langle \va, \vb \rangle \ge \epsDot $, where both 
$\epsilon_0, \epsDot > 0 $.
Then the product of all terms of the sequence converges to a rank-1 matrix $\bar C$:
\begin{equation}
    \bar C := \lim_{n\to \infty } \prod_{i=1}^n \frac{A_i}{|| \prod_{i=1}^n A_i ||}
    =  \bar \vc \bm{\gamma}^T  \,.
\end{equation}
\end{theorem}
\cite{hajnal1976productsNNM,friedland2006convergence} proved a similar result for squared matrices.
In appendix \ref{appendix:proof}, we provide a rigorous proof of the theorem using the cosine similarity.

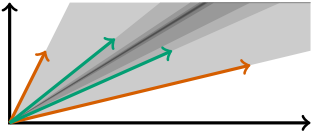
\begin{figure}[t]
    \centering
    \definecolor{cb1}{RGB}{213, 94, 0}
\definecolor{cb2}{RGB}{2, 158, 115}

\begin{tikzpicture}[very thick,scale=1.6]
    \pgfmathsetmacro{\h}{1}
    \pgfmathsetmacro{\w}{2.5}
    \coordinate (a1) at (\w * 0.2, \h);

    \draw [<->] (0,\h) node (yaxis) {}
        |- (\w,0) node (xaxis) [right] {};
    \fill[fill=black!20] (0,0) -- (a1) -- (\w, \h) -- (\w, 0.6) coordinate (b1) -- (0, 0);
    \fill[fill=black!30] (0,0) -- (\w * 0.5, \h) coordinate (a2) -- (\w, \h) -- (\w * 0.9, \h)  coordinate(b2) -- (0, 0) ;
    \fill[fill=black!40] (0,0) -- (\w * 0.6, \h) coordinate (a3) -- (\w, \h) -- (\w * 0.8, \h)  coordinate(b3) -- (0, 0) ;
    \fill[fill=black!50] (0,0) -- (\w * 0.63, \h) coordinate (a4) -- (\w, \h) -- (\w * 0.68, \h)  coordinate(b4) -- (0, 0) ;
    \fill[fill=black!60] (0,0) -- (\w * 0.64, \h) coordinate (a4) -- (\w, \h) -- (\w * 0.66, \h)  coordinate(b4) -- (0, 0) ;
    \fill[fill=black!70] (0,0) -- (\w * 0.645, \h) coordinate (a4) -- (\w, \h) -- (\w * 0.655, \h)  coordinate(b4) -- (0, 0) ;
    \draw[->,color=cb1] (0, 0) -- ($0.6*(a1)$);
    \draw[->,color=cb1] (0, 0) -- ($0.8*(b1)$);
    \draw[->,color=cb2] (0, 0) -- ($0.7*(a2)$);
    \draw[->,color=cb2] (0, 0) -- ($0.6*(b2)$);

\end{tikzpicture}
    \caption{
        The positive column vectors $\va_1, \va_2$ of matrix $A_1$ (orange) form a cone.
        The resulting columns of $A_1 A_2$ (green) are contained in the cone as they are
        positive linear combinations of $\va_1, \va_2$.
        At each iteration, the cone shrinks.
    }
    \label{fig:carton_cone}
\end{figure}

The geometric intuition of the proof is depicted in Figure \ref{fig:carton_cone}.
The column vectors of the first matrix are all non-negative and therefore in the positive quadrant.
For the matrix multiplication $A_i A_j$, observe that $A_i \va_k$ is a non-negative linear combination of the column vectors of $A_i$,
where $\va_k$ is the $k$-th column vector $A_{j_\idx{:k}}$.
The result will remain in the convex cone of the column vectors of $A_i$.
The conditions stated in the theorem ensure that the cone shrinks with every iteration and it converges towards a single vector.
In the appendix \ref{appendix:simulation_convergence}, we simulate different matrix properties and find non-negative matrices to converge exponentially fast.

The column vectors of a rank-1 matrix are linearly dependent $C\!=\!\vc\bm{\gamma}^T$. A rank-1 matrix $C$ always gives the same direction of $\vc$: $C Z^+_{k+1}\!=\!\vc\bm{\gamma}^T Z^+_{k+1}\!=\!\vc \bm{\lambda}^T $ and for any vector $\vv$: $ C Z^+_{k+1} \vv = \vc \bm{\lambda}^T \vv = t \vc $, where $t \in \R$.
For a finite number of matrices $C_k = \prod^k_l Z^+_l$, $C_k$ might resemble a rank-1 matrix up to floating-point imprecision
or $C_k Z^+_{k+1}$ might still be able to alter the direction. In any case, the influence of later matrices decreases.

The $Z^+$ matrices of dense layers fulfill the conditions of theorem 1.
Convolutions can be written as matrix multiplications.
For 1x1 convolutions, the kernels do not overlap and the row vectors corresponding to each location are orthogonal.
In this case, the convergence  happens only locally per feature map location. For convolutions with overlapping kernels, the global convergence is slower than for dense layers. In a ResNet-50 where the last convolutional stack has a size of (7x7), the overlapping of multiple (3x3) convolutions still induces a considerable global convergence (see \LRPcmp{} on ResNet-50 in section \ref{sec:lrp_cmp_paragraph}).

If an attribution method converges, the contributions of the layers shrink by depth. In the worst-case scenario, when converged up to floating-point imprecision,
the last layer can only change the scaling of the saliency map.
However, the last layer is responsible for the network's final prediction.

\subsection{Modified BP algorithms}

\paragraph{\zLRP}
The \zLRP \ rule of Layer-wise Relevance Propagation modifies the backpropagation rule as follows:
\newcommand{\rzlrp}[1]{r_{#1}^{z-\text{LRP}}}
\begin{equation}
\begin{split}
    \rzlrp{l}(\vx)
        &= Z_l \cdot \rzlrp{l+1}(\vx), \\
        \text{where}\, Z_{l} &= \left(
                \frac{w_{ij} \vh_{l_\idx{j}}}
                     {\sum_k w_{ik} \vh_{l_\idx{k}}}
            \right)_\idx{ij}^T.
\end{split}
\end{equation}
If only max-pooling, linear layers, and ReLU activations are used, it was shown that \zLRP \ corresponds to gradient$\odot$input, i.e. $\rzlrp{0}(\vx) = \vx \odot \frac{\partial f(\vx)}{\partial \vx}$ \citep{shrikumar2016lrp,kindermans2016investigating,ancona2017unified_view}.
\zLRP \ can be considered a gradient-based and not a modified BP method. The gradient is not converging to a rank-1 matrix and therefore gradient$\odot$input is also not converging.

\begin{figure*}[t]
    \input{export_defs/pattern_attr_s1_s1_prod.tex}
    \centering

    \begin{subfigure}[t]{0.33\textwidth}
        \includegraphics[width=0.9\textwidth]{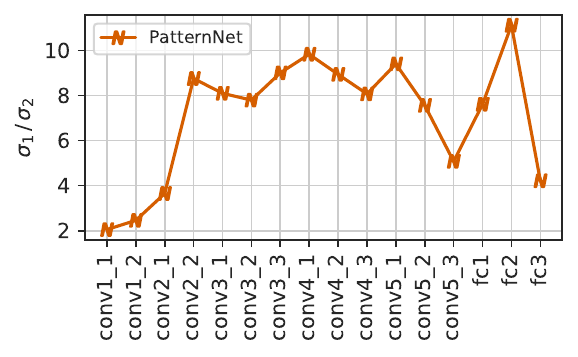}
         \caption{PatternNet}
         \label{fig:pattern_net_s1_s2_ratio}
    \end{subfigure}
    \begin{subfigure}[t]{0.33\textwidth}
        \includegraphics[width=0.9\linewidth]{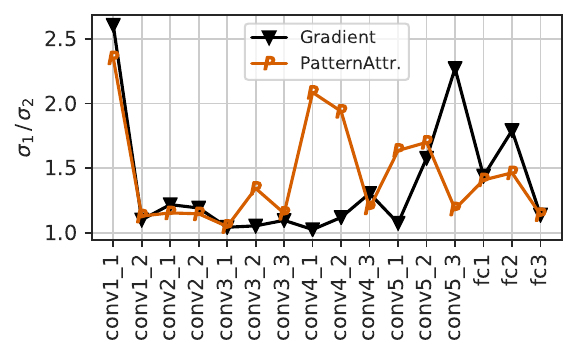}
        \caption{$W = U_l \Sigma_l V_l$}
    \end{subfigure}
    \begin{subfigure}[t]{0.33\textwidth}
        \includegraphics[width=0.9\linewidth]{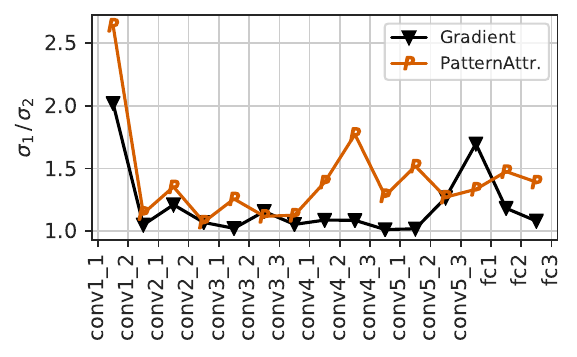}
         \caption{$T_l = \sqrt{\Sigma_l} V_l U_{l+1} \sqrt{\Sigma_{l+1}}$ }
    \end{subfigure}
    \caption{PatternNet \& PatternAttr.: \textbf{(a)(b)} Ratio between the first and second singular value $\sigma_1 / \sigma_2$ for $A_l, W_l,$ and $A_l\odot W_l$. \textbf{(c)} $\sigma_1 / \sigma_2$ of inter-layer derivation matrices.
    For (b) (c), we sliced the 3x3 convolutional kernels to 1x1 kernels.
    }
    \label{fig:pattern_attr_s1_s2_ratio}
\end{figure*}

\paragraph{\abLRP} separates the positive and negative influences:
\newcommand{\rab}[1]{r_{#1}^{\alpha\beta}}
\begin{equation}
    \rab{l}(\vx) = \left(\alpha {Z}_l^+ - \beta Z_l^-\right) \rab{l+1}(\vx),
\end{equation}
where $Z_l^+$ and $Z_l^-$ correspond to the positive and negative entries of the matrix $Z$.
\cite{bach2015lrp} propose to weight positives more: $\alpha \ge 1$ and $\alpha - \beta = 1 $. For \alphaBetaLRP{1}{0}, this rule corresponds to the $z^+$-rule, which converges. For $\alpha > 1$ and $\beta > 0$, the matrix $Z_l = \alpha {Z}_l^+ - \beta Z_l^- $ can contain negative entries. Our empirical results show that \abLRP{} still converges for the most commonly used parameters $\alpha=2, \beta=1$ and even for a higher $\alpha = 5$ it converges considerable on the ResNet-50.

\paragraph{Deep Taylor Decomposition} uses the $z^+$-rule if the input to a
convolutional or dense layer is in $[0, \infty]$, i.e. if the layer follows
a ReLU activation.  For inputs in $\R$, DTD also proposed the $w^2$-rule and the so-call $w^\mathcal{B}$ rule for bounded inputs. Both rules were specifically designed to produce non-negative outputs. Theorem \ref{Thm1} applies and DTD   converges to a rank-1 matrix necessarily.

\paragraph{PatternNet \& PatternAttribution} takes into account that the input $\vh_l$ contains noise. If $\vd_l$ corresponds to the noise and $\vs_l$ to the signal, then $\vh_l = \vs_l + \vd_l$. To assign the relevance towards the signal direction, it is estimated using the following equation:
\begin{equation}\label{eq:pattern_computed}
\va_i = \frac{\cov[\vh]\vw_i}{
        \vw^T_i \cov[\vh] \vw_i},
\end{equation}
where $a_i$ is the estimated signal direction for the $i-th$ neuron
with input $\vh$ and weight vector $\vw_i = W_\idx{i:}$.
PatternNet is designed to recover the relevant signal in the data.
Let $A_{l\idx{i:}} = \va_i$ be the corresponding signal matrix to the weight matrix $W_l$, the rule for PatternNet is:
\newcommand{\rPN}[1]{r_{#1}^\text{PN}}
\begin{equation}
        \rPN{l}(\vx) = A_l^T \cdot \rPN{l+1}(\vx),
\end{equation}
PatternNet is also prone to converge to a rank-1 matrix.
To recover the relevant signal, it might be even desired to converge to the a single direction -- the signal direction.

The convergence of PatternNet follows from the computation of the pattern vectors $\va_i$ in \eqref{eq:pattern_computed}. It is similar to a single step of the power iteration method $\vv_{k+1} = C \vv_k / \norm{C \vv_k}$.
In appendix \ref{appendix:patternattr}, we provide details on the relationship to power iteration and also derive \eqref{eq:pattern_computed} from the equation given in \citep{kindermans2018learning}.
The power iteration method converges to the eigenvector with the largest eigenvalue exponentially fast.

All column vectors in $A_\idx{i:} = \va_i$ underwent a single step
of the power iteration and therefore tend to point towards the first eigenvector of $\cov[\vh]$.
This can also be verified empirically: the ratio of the first and second singular value $\sigma_1(A) / \sigma_2(A) > 6$ for almost all the VGG-16 patterns (see Figure \ref{fig:pattern_net_s1_s2_ratio}), indicating a strong convergence of the matrix chain towards a single direction.

The findings from PatternNet are hard to transfer to PatternAttribution.
The rule for PatternAttribution uses the Hadamard product of $A_l$ and $W_l$:
\newcommand{\rPA}[1]{r_{#1}^\text{PA}}
\begin{equation}
        \rPA{l}(\vx) = \left(W_l\odot A_l\right)^T \cdot \rPA{l+1}(\vx),
\end{equation}
The Hadamard product complicates any analytic argument using the properties of
$A_l$ or $W_l$. The theoretical results available \citep{ando1987singularHadamard,zhan1997inequalities} did not
allow us to show that PatternAttribution converges to a rank-1 matrix necessarily.

We provide a mix of theoretical and empirical insights on why it converges.
The conditions of convergence can be studied well on the singular value decomposition: $(W_l \odot A_l)^T = U_l \Sigma_l V_l$. \todo{R4: $U \Sigma V$ not defined}
Loosely speaking, the matrix chain will converge to a rank-1 matrix if the first $\sigma_1$ and second  $\sigma_2$ singular values in $\Sigma_l$ differ and if $V_l$ and $U_{l+1}$ are aligned such that higher singular values of $\Sigma_l$ and $\Sigma_{l+1}$ are multiplied together such that the ratio $\sigma_1 / \sigma_2$ grows.

To see how well the per layer matrices align, we look at the inter-layer chain members: $T_l = \sqrt{\Sigma_l} V_l U_{l+1} \sqrt{\Sigma_{l+1}}$. In Figure \ref{fig:pattern_attr_s1_s2_ratio}, we display the ratio between the first and second singular values $\sigma_1(T_l) / \sigma_2(T_l)$. For $W \odot A$, the first singular value is considerably larger than for the plain weights $W$. Interestingly, the singular value ratio of inter-layer matrices shrinks for the plain $W$ matrix. Whereas for PatternAttribution,
the ratio increases for some layers indicating that the Hadamard product leads to more alignment of the matrices.

\paragraph{DeepLIFT} is the only tested modified BP method which does not converge to a rank-1 matrix. It is an extension of the backpropagation algorithm to finite differences:
\begin{equation}
     \frac{f(\vx) - f(\vx^0)}{\vx - \vx^0}
\end{equation}
For the gradient, one would take the limit $\vx^0 \to \vx$. DeepLIFT uses  a so-called reference point for $\vx^0$ instead, such as zeros or for images a blurred version of $\vx$.
The finite differences are backpropagated, similar to infinitesimal differences.
The final relevance is the difference in the $k$-th logit: $ r^{DL}_l(\vx) = f_k(x) -  f_k(x^0)$.

Additionally to the vanilla gradient, DeepLIFT separates positive and negative contributions. For ReLU activations, DeepLIFT uses either the RevealCancel or the Rescale rule. Please refer to \citep{shrikumar2017deeplift} for a description. The rule for linear layers is most interesting because it is the reason why DeepLIFT does not converge:
\begin{equation}
\label{eq:deeplift_linear}
\begin{split}
    r^{DL+}_l(\vx, \vx^0) = & M^T_{> 0} \odot \left(
        W_l^{+^T} r^{DL+}_{l+1}(\vx, \vx^0) \right. \\
    & \left. +  W_l^{-^T}  r^{DL-}_{l+1}(\vx, \vx^0) \vphantom{W_l^{-^T}} \right)
\end{split}
\end{equation}
where the mask $M_{> 0}$  selects the weight rows corresponding to positive deltas
($0 < \Delta \vh_l = \vh_l - \vh_l^0 $). For negative relevance $r^{DL-}_l$, the rule is defined analogously.
An interesting property of the rule (\ref{eq:deeplift_linear}) is that negative and positive relevance can influence each other.

If the intermixing is removed by only considering $W^+$ for the positive rule and $W^-$ for the negative rule, the two matrix chains become decoupled and converge. For the positive chain, this is clear. For the negative chain,
observe that the multiplication of two non-positive matrices gives a non-negative matrix.
Non-positive vectors $\vb, \vc$ have an angle $\le 90^{\circ}$ and
$\vc^T \vb = \norm{\vc} \norm{\vb} \cos(\vc, \vb) \ge 0$.
In the evaluation, we included this variant as \emph{DeepLIFT Ablation}, and as predicted by the theory, it converges.

\paragraph{Guided BP \& Deconv \& RectGrad} apply an additional ReLU to the gradient and it was shown  to be invariant to the randomization of later layers previously in \cite{adebayo2018sanity} and analyzed theoretically in \cite{nie2018theoretical}:
\newcommand{\rGBP}[1]{r_{#1}^{GBP}}
\begin{equation}
    \rGBP{l}(\vx) = W_l^T  \left[ M_{l} \rGBP{l+1}(\vx)  \right]^+.
\end{equation}
$M_{l} = \diag(\indicator_{h_1 > 0})$ denotes the gradient mask of the ReLU operation.
For Deconv, the mask of the forward ReLU is omitted, and the gradients are rectified directly.
RectGrad \cite{kim2019rectgrad} is related to GuidedBP and set the lowest $q$ percentile of the gradient to zero. As recommended
 in the paper, we used $q = 98$.

As a ReLU operation is applied to the gradient, the backpropagation is no longer a linear function.
The ReLU also results  in a different failure than before. \citep{nie2018theoretical} provides a theoretical analysis
for GuidedBP. Our results align with them.

\section{Evaluation}

\begin{figure*}[t]
    \centering
    \begin{subfigure}[t]{0.48\textwidth}
    \centering
    \includegraphics[width=0.60\linewidth]{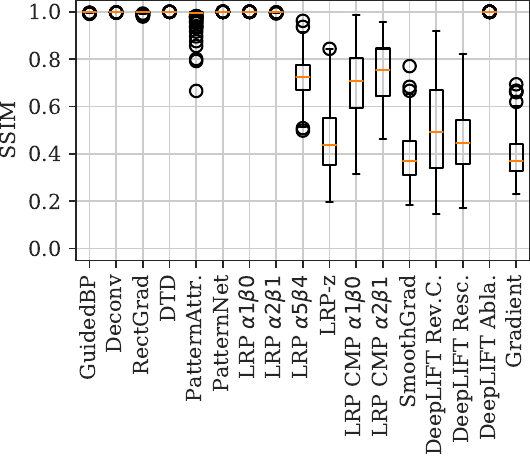}
        \caption{Random Logits}
    \label{fig:random_logits}
\end{subfigure}
    \begin{subfigure}[t]{0.48\textwidth}
    \centering
    \includegraphics[width=0.90\linewidth]{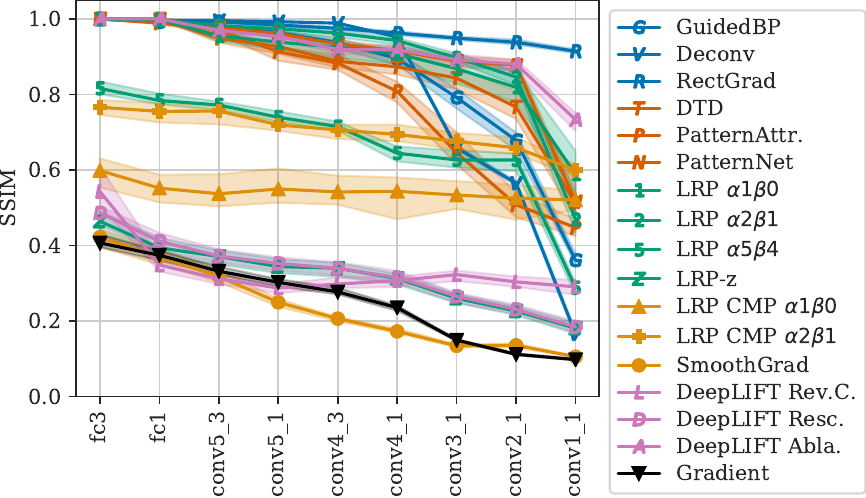}
        \caption{Parameter Randomization}
    \label{fig:ssim_random_params_vgg16}
\end{subfigure}
    \caption{%
        \textbf{(a)}
        SSIM between saliency maps explaining the ground-truth or a random logit.
        \textbf{(b)} The parameters of the VGG-16 are randomized, starting from the last to the first layer.  SSIM quantifies the difference to the saliency map from the original model.  Intervals show 99\% bootstrap confidences.
    }
\end{figure*}

\paragraph{Setup} We report results on a small network trained on CIFAR-10 (4x conv., 2x dense, see appendix \ref{appendix:cifar10}),
a VGG-16 \citep{simonyan2014vgg}, and ResNet-50 \citep{he2016resnet}. The last two are trained on the ImageNet dataset \cite{ILSVRC15}, the standard dataset to evaluate attribution methods.
The different networks cover different concepts: shallow vs. deep, forward vs. residual connections, multiple dense layers vs. a single one, using batch normalization.
All results were computed on 200 images from the validation set. To justify the sample size, we show bootstrap confidence intervals in Figure \ref{fig:ssim_random_params_vgg16} \citep{efron1979bootstrap}. We used the implementation from the \emph{innvestigate} and \emph{deeplift} package \citep{alber2019innvestigate,shrikumar2017deeplift} and added support for residual connections. The experiments were run on a single machine with two graphic cards and take about a day to complete.

\paragraph{Random Logit} We display the difference of saliency maps explaining the ground-truth and a random logit in Figure \ref{fig:random_logits}.
As the logit value is responsible for the predicted class, the saliency maps should change.
We use the SSIM metric \citep{wang2004ssim} as in \cite{adebayo2018sanity}.

\paragraph{Sanity Check}
We followed \cite{adebayo2018sanity} and randomized the parameters starting from the last layer to the first layer.
For DTD and \aIbOLRP, randomizing the last layer flips the sign of the saliency map sometimes. We, therefore, compute
the SSIM also between the inverted saliency map and report the maximum.
In Figure \ref{fig:ssim_random_params_vgg16}, we report the SSIM between the saliency maps
(see also Figure \ref{fig:figure1_sanity_checks} and appendix \ref{appendix:sanity_checks}).\footnotemark{}

\footnotetext{For GuidedBP, we report different saliency maps than shown in Figure 2 of \cite{adebayo2018sanity}. We were able to confirm a bug in their implementation, resulting in saliency maps of GuidedBP and Guided-GradCAM to remain identical for early layers.}

\newcommand{\cossimHeight}{3.1cm}
\newcommand{\cosHistHeight}{2.8cm}
\begin{figure*}[t]
    \centering
    \begin{subfigure}[t]{1\textwidth}
        \includegraphics[width=\textwidth,height=\cossimHeight]{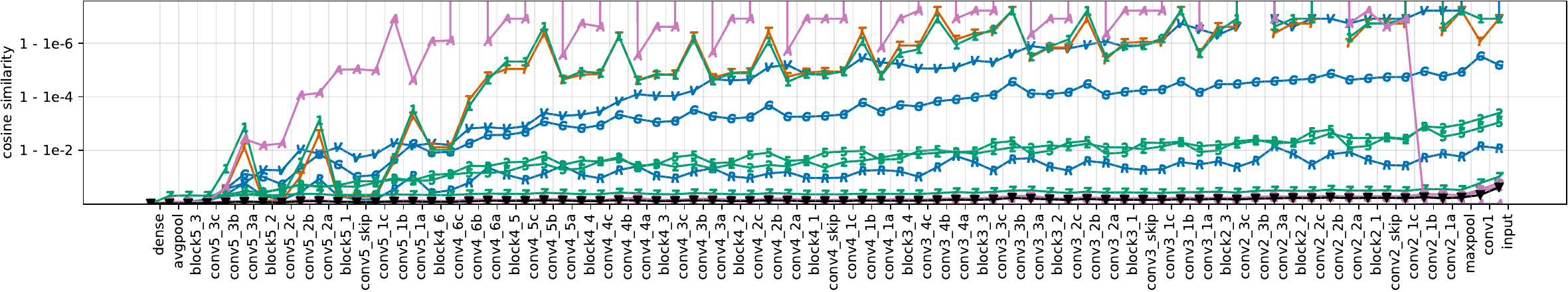}
        \caption{ResNet-50}
        \label{fig:csc_resnet50}
    \end{subfigure}

    \begin{subfigure}[t]{0.4\textwidth}
        \includegraphics[height=\cossimHeight]{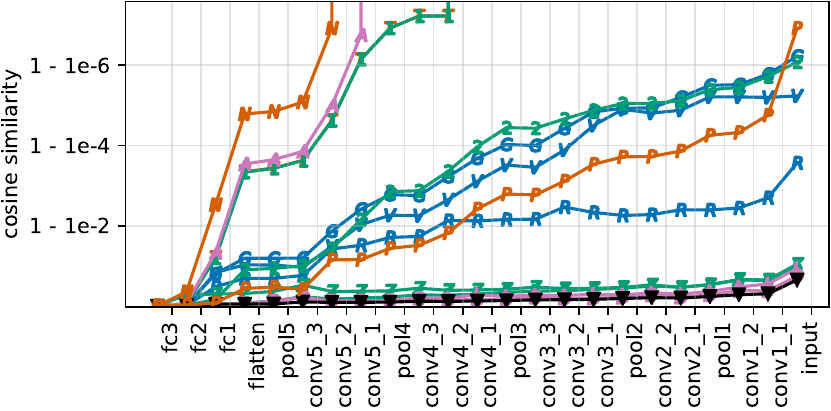}
        \caption{VGG-16 (logarithmic)}
        \label{fig:csc_vgg16_log}
    \end{subfigure}
    \begin{subfigure}[t]{0.4\textwidth}
        \includegraphics[height=\cossimHeight]{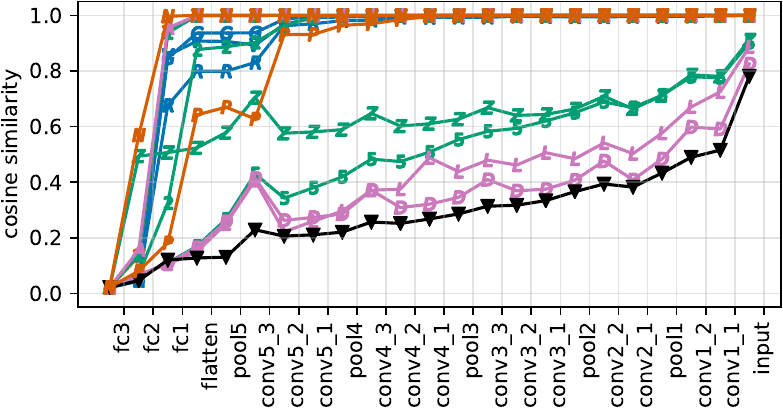}

        \caption{VGG-16 (linear)}
        \label{fig:csc_vgg16_linear}
    \end{subfigure}
    \begin{subfigure}[t]{0.18\textwidth}
        \raisebox{0.0cm}{
            \includegraphics[height=\cossimHeight]{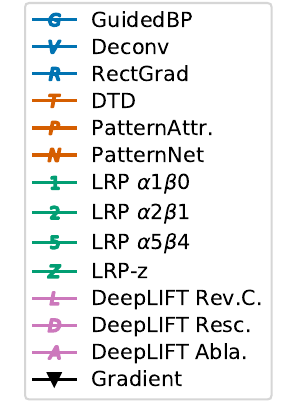}
        }
    \end{subfigure}

    \begin{subfigure}[t]{0.25\textwidth}
        \includegraphics[height=\cossimHeight]{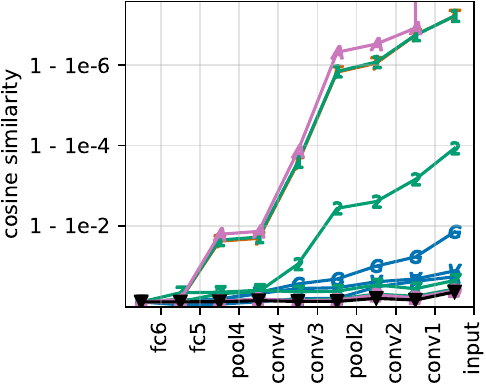}
        \caption{CIFAR-10}
        \label{fig:csc_cifar10}
    \end{subfigure}
    \begin{subfigure}[t]{0.18\textwidth}
        \includegraphics[height=\cosHistHeight]{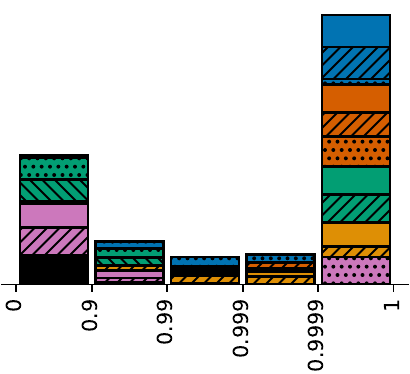}
        \caption{VGG-16}
        \label{fig:csc_hist_vgg16}
    \end{subfigure}
    \begin{subfigure}[t]{0.18\textwidth}
        \includegraphics[height=\cosHistHeight]{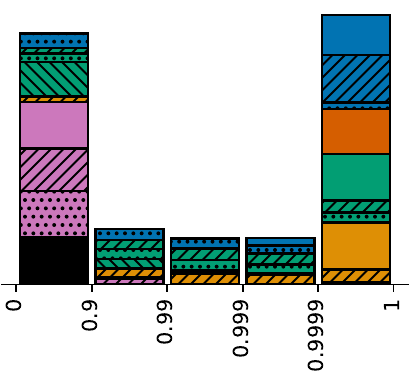}
        \caption{ResNet-50}
        \label{fig:csc_hist_resnet50}
    \end{subfigure}
    \begin{subfigure}[t]{0.18\textwidth}

        \includegraphics[height=\cosHistHeight]{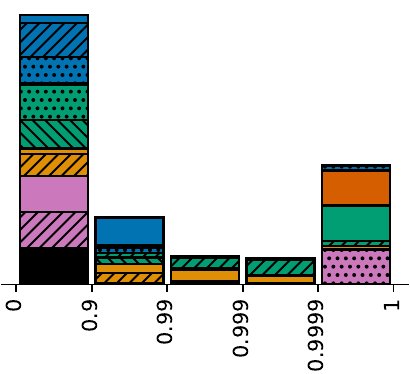}

        \caption{CIFAR10}
        \label{fig:csc_hist_cifar}
    \end{subfigure}
    \begin{subfigure}[t]{0.15\textwidth}
        \raisebox{0cm}{\includegraphics[height=\cossimHeight]{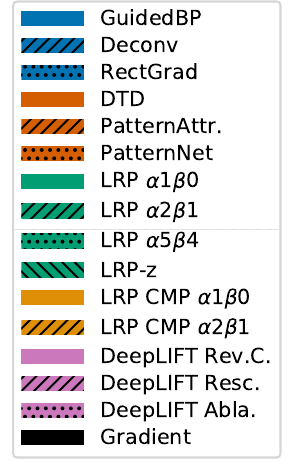}}
    \end{subfigure}

    \hfill

    \caption{\textbf{(a)-(d)} Median of the cosine similarity convergence (CSC) per layer between relevance vectors obtained from randomizing the relevance vectors of the final layer. \textbf{(e)}-\textbf{(g)} histogram of the distribution of the CSC after the first layer.
    }
    \label{fig:csc_large_figure}
\end{figure*}

\paragraph{Cosine Similarity Convergence Metric (CSC)}
Instead of randomizing the parameters, we randomize the backpropagated relevance vectors directly. We select layer $k$ and set the corresponding relevance to $r_{k}(\vx) := \vv_1 $ where $\vv_1  \sim \mathcal{N}(0, I)$ and then backpropagate it as before.
For example, for the gradient, we would do: $\frac{\partial h_k}{\partial h_1}\frac{\partial f(x)}{\partial h_k} := \frac{\partial h_k}{\partial h_1}\vv_1$. We use the notation $ r_l(\vx | r_k\!:\!\shorteq \vv_1)$ to describe the relevance $r_l$ at layer $l$ when the relevance of layer $k$ is set to $\vv_1$.

Using two random relevance vectors $\vv_1, \vv_2  \sim \mathcal{N}(0, I)$, we measure the convergence using the cosine similarity.  A rank-1 matrix $C\!=\!\vc\vgamma^T $ always yields the same direction:  $ C \vv\!=\vc\vgamma^T\!=\!\lambda \vc$.
\newcommand{\relv}[1]{r_l\left(\vx|r_k{:\shorteq}\vv_#1\right)}
If the matrix chain converges, the backpropagated relevance vectors of $\vv_1, \vv_2 $ will align more and more. We quantify their alignment using the cosine similarity $\scos(\relv{1},\relv{2}))$
where $\scos(\va, \vb) = \va^T \vb \ / \ \left(\norm{\va}\cdot\norm{\vb}\right)$.

Suppose the relevance matrix chain would converge to a rank-1 matrix perfectly, than we have for both $\vv_1, \vv_2$: $r_l(\vx|r_k\shorteq\vv_i) = C \vv_i = c \bm{\gamma}^T \vv_i = \lambda_i  \vc$ where $\lambda_i = \bm{\gamma}^T \vv_i $  and their cosine similarity will be one. The opposite direction is also true.
If $ C $ has shape $n\times m$ with $n \le m$ and if for $n$ linearly independent vectors $\vv_i$, the cosine similarity $\scos( C \vv_i,  C \vv_j) = 1$, then $ C$ is a rank-1 matrix.

An alternative way to measure convergence would have been to construct the derivation matrix $C_k = \prod^k_{l=1} Z_l$ and measure the ratio $\sigma_1(C_k) / \sigma_2(C_k)$ of the first to the second-largest singular value of $C_k$. Although this approach is well motivated theoretically, it has some performance downsides. $C_k$ would be large and computing the singular values costly.

We use five different random vectors per sample -- in total 1000 convergence paths. As the vectors are sampled randomly, it is unlikely to miss a region of non-convergence \cite{bergstra2012random}.

For convolution layers, we compute the cosine similarity per feature map location. For a shape of $(h, w, c)$, we obtain $h\cdot w$ values.
The jump in cosine similarity for the input is a result of the input's low dimension of 3 channels.
In Figure \ref{fig:csc_large_figure}, we plot the median cosine similarity for
different networks and attribution methods (see appendix \ref{appendix:csc_figures} for
additional Figures). We also report the histogram of the CSC at the first convolutional layer in Figures \ref{fig:csc_hist_vgg16}-\ref{fig:csc_hist_cifar}.

\section{Results}
Our random logit analysis reveals that converging methods produce almost identical saliency maps, independently of the output logit (SSIM very close to 1). The rest of the field (SSIM between 0.4 and 0.8) produces saliency maps different from the ground-truth logit's map (see Figure \ref{fig:random_logits}).

We observe the same distribution in the sanity check results (see Figure \ref{fig:ssim_random_params_vgg16}). One group of methods produces similar saliency maps even when convolutional layers are randomized (SSIM close to 1). Again, the rest of the field is sensitive to parameter randomization. The same clustering
can be observed for ResNet-50 (appendix \ref{appendix:result_resnet}, Figure \ref{fig:sanity_checks_resnet50}).

Our CSC analysis confirms that random relevance vectors align throughout the backpropagation steps (see Figure \ref{fig:csc_large_figure}). Except for \zLRP{} and DeepLIFT, all methods show convergence up to at least 0.99 cosine similarity.
\alphaBetaLRP{5}{4}{} converges less strongly for VGG-16. Among the converging methods, the rate of convergence varies. \aIbOLRP{}, PatternNet, the ablation of DeepLIFT converges fastest. PatternAttribution has a slower convergence rate -- still exponential.
For DeepLIFT Ablation, numerical instabilities
result in a cosine similarity of 0 for the first layers of
the ResNet-50. Even on the small 6-layer network, the median CSC is greater than 1-1e-6 for \aIbOLRP{} (see Figure \ref{fig:csc_cifar10}).

\section{Discussion}

When many modified BP methods do not explain the network faithfully, why was this not widely noticed before?
First, it is easy to blame the network for unreasonable explanations -- no ground truth exists.
Second, MNIST, CIFAR, and ImageNet contain only a single object class per image -- not revealing the class insensitivity.
Finally, it might not be too problematic for some applications if the saliency maps are independent of the later network's layers.
For example, to explain Alzheimer's disease \citep{bohle2019lrp}, local low-level  features are sufficient as they are predictive for the disease and the data lacks conflicting evidences (i.e. the whole brain is affected).

When noticed, different ways to address the issue were proposed and an improved class sensitivity was reported \citep{kohlbrenner2019bestLRP,gu2018understanding,zhang2018excitationbp}.
We find that the underlying convergence problem remains unchanged and discuss the methods below.

\begin{figure}[t]
    \centering
    \begin{subfigure}[t]{\linewidth}
        \includegraphics[width=0.9\linewidth]{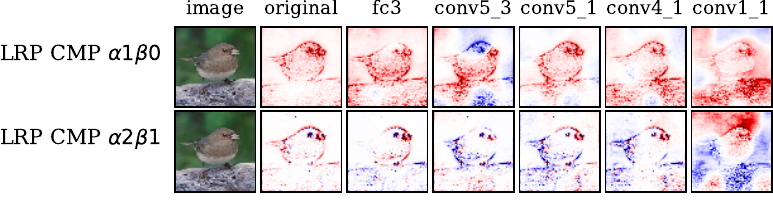}
        \caption{VGG-16}
        \label{fig:sanity_check_lrp_cmp_vgg}
     \end{subfigure}
     \begin{subfigure}[t]{\linewidth}
        \includegraphics[width=0.9\linewidth]{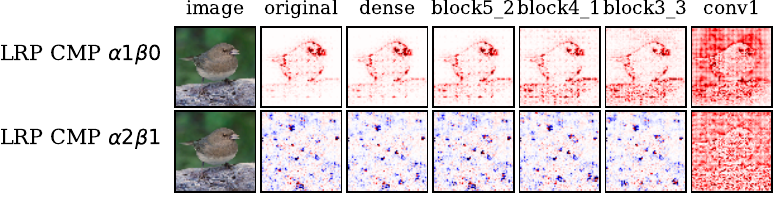}
        \caption{ResNet-50}
        \label{fig:sanity_check_lrp_cmp_resnet}
     \end{subfigure}
     \begin{subfigure}[t]{\linewidth}
        \vspace{-0.5cm}
        \includegraphics[width=0.68\linewidth]{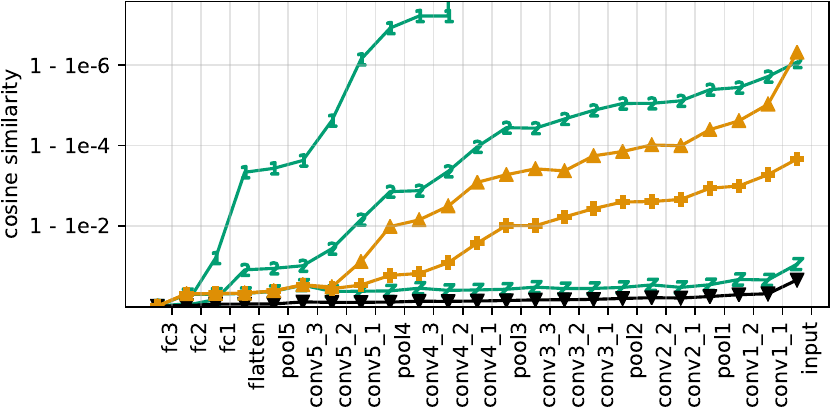}
        \raisebox{0.5cm}{\includegraphics[width=0.3\linewidth]{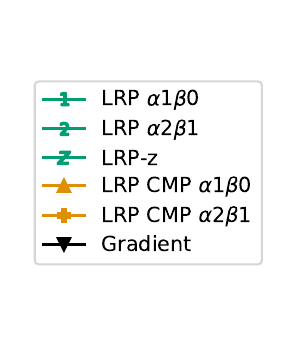}}
        \caption{VGG-16}
        \label{fig:cosine_similiarity_lrp_cmp_vgg}
     \end{subfigure}
    \caption{\textbf{(a-b)} Sanity checks and \textbf{(c)} CSC for \LRPcmp{}.}
    \label{fig:sanity_check_lrp_cmp}
    \vspace{-0.5cm}
\end{figure}

\paragraph{\LRPcmp} \label{sec:lrp_cmp_paragraph}
\citep{kohlbrenner2019bestLRP,lapuschkin2017age_gender} use \zLRP \ for the final dense layers and \abLRP \ for the convolutional layer. We report results for $\alpha=1, 2$ as in \citep{kohlbrenner2019bestLRP} in Figure \ref{fig:sanity_check_lrp_cmp_vgg}.

For VGG-16, the saliency maps change when the network parameters are randomized. However, structurally, the underlying image structure seems to be scaled only locally (see Figure \ref{fig:sanity_check_lrp_cmp_vgg}). Inspecting the CSC path of the two \LRPcmp\ variants in Figure \ref{fig:cosine_similiarity_lrp_cmp_vgg}, we can see why. For dense layers, both methods do not converge as \zLRP \ is used, but the convergence start when \abLRP \ is applied.
The relevance vectors of the dense layer can change the coarse local scaling. However, they cannot alter the direction of the relevance vectors of earlier layers to highlight different details.

In the backward-pass of the ResNet-50, the global\hyp{}averaging layer assigns the identical gradient vector to each location of the last convolutional layer. Furthermore, the later convolutional layers operate on (7x7), where even a few 3x3 convolutions have a dense field-of-view. \LRPcmp{} does not resolve the global convergence for the ResNet-50.

\paragraph{Contrastive LRP} \cite{gu2018understanding} noted the lack of class sensitivity and proposed to increase it by subtracting two saliency maps. The first saliency map explains only the logit $\vy_k = \vy \odot \vm_k$, where $\vm_k$ is a one-hot vector
and the second explains the opposite $\vy_{\neg k}= \vy \odot (1 - \vm_k)$:
\begin{equation}
\begin{split}
    \max(0, & n(\rzplus{\vx}(\vx|r_{\text{logits}}{:\shorteq}\vy_k)) \\
     & - n(\rzplus{\vx}(\vx|r_{\text{logits}}{:\!\shorteq}\, \vy_{\neg k}))
\end{split}
\end{equation}
$n( . )$ normalizes each saliency map by its sum. The results of Contrastive LRP are similar to Figre \ref{fig:figure1_diff_cat_dog}, no $\max$ is applied. The underlying convergence problem is not resolved.

\paragraph{Contrastive Excitation BP}
The lack of class sensitivity of the $z^+$-rule was noted in \cite{zhang2018excitationbp} and to increase it, they proposed to change the backpropagation rule of the final fully-connected layer to:
\begin{equation}
    r^\text{cEBP}_{\text{final fc}}(x) = (Z^+_\text{final fc} - N^+_\text{final fc}) \vm_k,
\end{equation}
where $\vm_k$ is a one-hot vector selecting the explained class.
The added $N^+_\text{final fc}$ is computed as the $Z^+_\text{final fc}$ but on the negative weights $-W_\text{final fc}$.  Note that the combination of the two matrices introduces negative entries. Class sensitivity is increased. It does also not resolve the underlying convergence problem. If, for example, more fully-connected layers would be used, the saliency maps would become globally class insensitive again.

\paragraph{Texture vs. Contours} \cite{Geirhos2018ImageNettrainedCA} found that
deep convolutional networks are more sensitive towards texture and not the shape
of the object. For example, the shape of a cat filled with an elephant texture
will be wrongly classified as an elephant. However, modified BP methods
highlight the contours of objects rather.

\paragraph{Recurrent Neural Networks}
Modified BP methods are focused on convolutional neural networks and are mostly applied on vision tasks. The \emph{innvestigate} package does not yet support recurrent models.
To our knowledge, \cite{arras2017explaining} is the only work that applied modified BP rules to RNNs (\zLRP{} for LSTMs). Training, and applying modified backpropagation rules to RNNs, involves unrolling the network, essentially transforming it to a feed-forward architecture. Due to our theoretical results, modified BP rules that yield positive relevance matrices (e.g. $z^+$-rule) will converge. However, further work would be needed to measure how RNN architectures (LSTM, GRU) differ in their specific convergence behavior.

\paragraph{Not Converging Attribution Methods}
Besides modified BP attribution methods, there also exist gradient averaging and black-box methods.
\emph{SmoothGrad} \citep{smoothgrad} and \emph{Integrated Gradients} \citep{sundararajan2017axiomatic} average the gradient.
\emph{CAM} and \emph{Grad-CAM} \citep{cam,selvaraju2017gradcam}
determine important areas by the activation of the last convolutional layer.
Black-box attribution methods only modify the model's input but do not rely on the gradient or other model internals.
The most prominent black-box methods are \emph{Occlusion}, \emph{LIME}, \emph{SHAP} \citep{zeiler2014deconvnet,ribeiro2016lime,lundberg2017shap}.
\emph{IBA} \citep{schulz2020restricting} applies an information bottleneck to remove unimportant information.
\emph{TCAV} \cite{kim2018interpretability} explains models using higher-level concepts.

All here mentioned attribution methods do \emph{not} converge, as they either rely on the gradient or treat the model as black-box. Only when the BP algorithm is modified, the convergence problem can occur. The here mentioned algorithms might still suffer from other limitations.

\paragraph{Limitations} Also, we tried to include most modified BP attribution
methods, we left some out for our evaluation
\cite{nam2019rap,wang2019bias,huber2019selectivelrp}.
In our theoretical analysis of
PatternAttribution, we based our argument on why it converges on
empirical observations performed on a single set of pattern matrices.

\section{Related Work}

\paragraph{Limitations of attribution}
The limitations of explanation methods were studied before.
\cite{viering2019gradcamCase} alter the explanations of Grad-CAM arbitrarily by modifying the model architecture only slightly.
Similarly, \cite{slack2020fooling_lime} construct a biased classifier that can hide its biases from LIME and SHAP.
The theoretic analysis \citep{nie2018theoretical} indicates that
GuidedBP tends to reconstruct the input instead of explaining the network's decision.
\citep{adebayo2018sanity} showed GuidedBP to be independent of later layers' parameters.
\citep{atrey2020exploratory} tested saliency methods in a reinforcement learning setting.

\citep{kindermans2018learning} show that LRP, GuidedBP, and Deconv produce incorrect
explanations for linear models if the input contains noise. \cite{rieger2017separable,zhang2018excitationbp,gu2018understanding,kohlbrenner2019bestLRP,montavon2019overview,tsunakawa2019crp}
noted the class-insensitivity of different modified BP methods, but they rather proposed ways to improve the class sensitivity
than to provide correct reasons why modified BP methods are class insensitive.
Other than argued in \cite{gu2018understanding}, the class insensitivity is not
caused by missing ReLU masks and Pooling switches. To the best of our knowledge, we are the first to identify the reason why many modified BP methods do not explain the decision of deep neural networks faithfully.

\paragraph{Evaluation metrics for attribution}

As no ground-truth data exists for feature importance, different proxy tasks were proposed to measure the performance of attribution algorithms.
One approach is to test how much relevance falls into ground-truth bounding boxes \cite{schulz2020restricting,zhang2018excitationbp}.

The \emph{MoRF} and \emph{LeRF} evaluation removes the \emph{mo}st and \emph{le}ast relevant input features and measures the change in model performance \cite{samek2016evaluating}. The relevant image parts are masked usually to zero. On these modified samples, the model might not be
reliable. The \emph{ROAR} score improves it by retraining the model from scratch \citep{hooker2018roar}. While computationally expensive, it ensures the model performance does not drop due to out-of-distribution samples. The ROAR performance of Int.Grad. and GuidedBP is equally bad, worse than a random baseline (see Figre 4 in \citep{hooker2018roar}). Thus, ROAR does not separate converging from non-converging methods.

Our CSC measure has some similarities with the work \cite{balduzzi2017shattered}, which analyzes the effect of skip connections on the gradient. They measure the convergence between the gradient vector from different samples using the effective rank \citep{vershynin2012random}.
The CSC metric applies to modified BP methods and is an efficient tool to trace the degree of convergence.

A different approach to verify attribution methods is to measure how helpful they are for humans \cite{alqaraawi2020userstudy,doshi2017towards,lage2018human}.

\section{Conclusion}

In our paper, we analyzed modified BP methods, which aim to explain the predictions of deep neural networks.
Our analysis revealed that most of these attribution methods have theoretical properties contrary to their goal.
PatternAttribution and LRP cite Deep Taylor Decomposition as the theoretical motivation. In the light of our results, revisiting the theoretical derivation of Deep Taylor Decomposition may prove insightful. Our theoretical analysis stresses the importance of negative relevance values. A possible way to increase class-sensitivity and resolve the convergence problem could be to backpropagate negative relevance similar to DeepLIFT, the only method passing our test.

\section{Acknowledgements}

We are grateful to the comments by our reviewers, which help to improve the manuscript further. We thank Benjamin Wild and David Dormagen for stimulating discussions.  We also thank Avanti Shrikumar for answering our questions and helping us with the DeepLIFT implementation.  The comments by Agathe Balayn, Karl Schulz, and Julian Stastny improved the manuscript. 
Furthermore, we thank Günter Rote for highlighting an inaccuracy in Theorem 1, leading to its current revision and update.
A special thanks goes to the anonymous reviewer 1 of our paper \cite{schulz2020restricting}, who encouraged us to report results on the sanity checks --- the starting point of this paper.
The Elsa-Neumann-Scholarship by the state of Berlin supported LS.
We are also grateful to Nvidia for a Titan Xp and to ZEDAT for access their HPC system.

\bibliographystyle{icml2020}
\bibliography{references}

\onecolumn

\appendix

\section{Proof of Theorem 1}
\label{appendix:proof}
In \cite{friedland2006convergence} the theorem is proven for square matrices. In fact Theorem 1 can be deduced from this case by the following argument:

For a sequence of non-square matrices $(A_k)_{k\in \mathbb{N}}\in \mathbb{R}^{m_k\times l_k}$ of finite size $m_k,l_k\leq L$ we can always find a finite set of subsequent matrices that when multiplied together are a square matrix.
\begin{align}
    \underbrace{A_1\cdot...\cdot A_{n_1}}_{=:\bar{A}_1\in \mathbb{R}^{m\times m}}\cdot \underbrace{A_{n_1+1}\cdot...\cdot A_{n_2}}_{=:\bar{A}_2\in \mathbb{R}^{m\times m}}\cdot\underbrace{A_{n_2+1}\cdot...\cdot A_{n_3}}_{=:\bar{A}_3\in \mathbb{R}^{m\times m}}\cdot ...
\end{align}
The matrices $\bar{A}_1,\bar{A}_2,\bar{A}_3,...$ define a sequence of non-negative square matrices that fulfill the conditions in \cite{friedland2006convergence} and therefore converge to a rank-1 matrix.

We provide another proof using the cosine similarity to show
convergence.
First, we outline the conditions on the matrix sequence $A_n$. Then, we state the theorem again and sketch our proof to give the reader a better overview. Finally, we prove the theorem in 4 steps.

\paragraph{Conditions on $A_n$} The first obvious condition is that the $(A_n)_{n\in \mathbb{N}}$ is a sequence of non-negative matrices such that $A_i$, $A_{i+1}$ have the correct size to be multiplied together.
Secondly, as we calculate angles between column vector in our proof, no column of $A_n$ should be zero. The angle between a zero vector and any other vector is undefined.
Furthermore, we assume that the entries of $A_n$ cannot grow infinitely. They are bound such that $||\va||_1 \le L \in \R^+$ for all column vectors $\va$ of any $A_n$.
Finally, the size of $A_n$ should not increase infinitely, i.e. an upper bound on the size of the $A_i$'s exists such that $A_i\in \mathbb{R}^{m\times k}$ where $m,k\leq K$ for some $K\in \mathbb{N}$.

\textbf{Theorem 1.}
Let $A_1, A_2, A_3\dots$ be a sequence of non-negative matrices as described above. 
We require that every column vector $\va$ of $A_n$ has a norm $|| \va || \ge \epsilon_0$
and that infinite many matrices $A_i$ with $i \in I$ and $|I| = |\mathbb{N}|$ exists for which two column vectors have a dot product of at least $\epsilon_{\langle \cdot,  \cdot \rangle }$, i.e. 
$\langle \va, \vb \rangle \ge \epsDot $, where both 
$\epsilon_0, \epsDot > 0 $.
Then the product of all terms of the sequence converges to a rank-1 matrix $\bar C$:
\begin{equation}
    \bar C := \lim_{n\to \infty } \prod_{i=1}^n \frac{A_i}{|| \prod_{i=1}^n A_i ||}
    =  \bar \vc \bm{\gamma}^T  \,.
\end{equation}

\paragraph{Example} 
Infinite many matrices in $A_n$ must not be too orthogonal $\left( \langle \vv_i, \vv_j \rangle < \epsDot \right)$
or be too close to the zero vector $\left( |\vv_i| < \epsilon_0 \right)$.
Matrices of the following form are \emph{not in} the sequence $A_n$:
\begin{center}
    $\begin{pmatrix}
        \undermat{\text{arbitrary}}{\begin{matrix}
        & &  \\
        \vv_1&...&\vv_l\\
        & &
        \end{matrix}} &
        \undermat{\langle \vv_i, \vv_j \rangle < \epsDot
        }{\begin{matrix}
        & &  \\
        \vv_{l+1}&...&\vv_m\\
        & &
        \end{matrix}
        }
        & 
        \undermat{|\vv_i| < \epsilon_0 }{
            \begin{matrix}
            & &  \\
            \vv_{m+1}&...&\vv_n\\
            & &
            \end{matrix}
        }
    \end{pmatrix}$
\end{center}
up to ordering of the columns.

For example, this concrete example could only occur finite times:
\begin{center}
    $\begin{pmatrix}
        \begin{matrix}
        0& 1\\
        1& 1\\
        1& 1\\
        1& 1
        \end{matrix}& 
        \undermat{
            \langle \vv_3, \vv_4 \rangle < \epsDot 
        }{
            \begin{matrix}
            0.5& \epsDot \\
            0& 0\\
            1& 0\\
            0& 1
            \end{matrix}
        }
        \hspace{0.5cm}
        & 
        \undermat{||\vv_i || < \epsilon_0}
        {
            \begin{matrix}
            0& 0\\
            0& 0\\
            0& 0\\
            0& 0.5 \epsilon_0
            \end{matrix}
        }
    \end{pmatrix}.$
\end{center}

\vspace{1cm}
\paragraph{Proof sketch}
To show that $\prod_i^\infty A_i$ converges to a rank-1 matrix, we do the following steps:
\begin{enumerate}[label={\textbf{(\arabic*)}}]
    \item We define a sequence $s_n$ as the cosine of the maximum angle between the column vectors of $M_n := \prod^n_{i=1} A_i$.
    \item We show that the sequence $s_n$ is monotonic and bounded and therefore converging.
    \item  We introduce a complementary sequence $\varepsilon_n = 1 - s_n$ and show $s_{n+1} \ge s_n + \lambda \varepsilon_n$ with $\lambda > 0$
    \item  
    We assume that $\lim_{n\to\infty} s_n = 1-\varepsilon^* < 1$ and show that than $\lim_{n\to\infty} s_n \to \infty$ is diverging which is a contradiction and therefore $s_n \to 1$.
\end{enumerate}

\renewenvironment{proof}{{\textbf{{Proof}} \hspace{0.2cm}}}{\qed}

\begin{proof}
To simplify the proof, we assume that all matrices in the sequence $A_i$ are in $I$. We will show that any finite number of $(A_i)_{i\not \in I}$ can be added without changing the result.

\textbf{(1)} Let $M_n:=\prod_{i=1}^n A_i$ be the product of the matrices $A_1\cdot \hdots \cdot A_n$. We define a sequence on the angles of column vectors of $M_n$ using the cosine similarity. Let $\vv_1(n),...,\vv_{k(n)}(n)$ be the column vectors of $M_n$.
Note, the angles are well defined between the columns of $M_n$. The columns of $M_n$ cannot be a zero vector as we required $A_n$ to have no zero columns.
Let $s_n$ be the cosine of the maximal angle between
the columns of $M_n$:
\begin{align}
    s_n = \operatorname{min}_{i\neq j} \operatorname{\scos}(\vv_i(n),\vv_j(n)):= \min_{i,j} \frac{\langle \vv_i(n),\vv_j(n)\rangle}{\|\vv_i(n)\|\|\vv_j(n)\|},
\end{align}
where $\langle\cdot,\cdot\rangle$ denotes the dot product. We show that the maximal angle  converges to 0 as $\lim_{n\to \infty} s_n = 1$, which is equivalent to $M_n$ converging to a rank-1 matrix. In the following, we take a look at two consecutive elements of the sequence $s_n$ and check by how much the sequence increases.
\newcommand{\SumaSumb}{\left(\sum_i a_i\right)\left(\sum_i b_i \right)}

\textbf{(2)} We show that the sequence $s_n$ is monotonic and bounded and therefore converging.  Assume  $\va_{n+1}$ and $\vb_{n+1}$ are the two columns of $A_{n+1}$ which produce the columns $\vv_m(n+1)$ and $\vv_{m'}(n+1)$ of $M_{n+1}$ with the maximum angle:
\begin{align}
    s_{n+1}= \operatorname{\scos}(\vv_{m}(n+1),\vv_{m'}(n+1))=\operatorname{\scos}(M_n\va_{n+1},M_n\vb_{n+1}).
\end{align}
We also assume that $\|\vv_i(n)\|=1$ for all $i$, since the angle is independent of length.
To declutter notation, we write $\vv_i(n)=:\vv_i$, $\va_n=\va=(a_1,...,a_{k})^T$, $\vb_n=\vb=(b_1,...,b_{k})^T$

Substituting $M_n \va$ = $ \sum_i a_i \vv_i$ into the
the definition of the cosine similarity,
we show that $s_n$ is monotonic:
\begin{align}\label{eq:s_n+1}
    s_{n+1}=\frac{\sum_{ij} a_ib_j\langle \vv_i,\vv_j \rangle}{\|\sum_{i} a_i \vv_i\|\|\sum_{i} b_i \vv_i\|}
\end{align}
Using the triangle inequality  $\|\sum_i a_i \vv_i\| \leq \sum_i a_i \|\vv_i\|$ we get:
\begin{align}
    s_{n+1} \geq \frac{\sum_{ij} a_ib_j\langle \vv_i,\vv_j \rangle}{(\sum_{i} a_i\| \vv_i\|)(\sum_{i} b_i \|\vv_i\|)}
\end{align}
As we assumed that the $\norm{\vv_i} = 1$, we know that $ \langle \vv_i,\vv_j \rangle = \scos(\vv_i, \vv_j)$ which must be greater than the smallest cosine similarity $s_n$:
\begin{align}\label{eq:s_n_is_monotonic}
    s_{n+1} \geq \frac{\sum_{ij} a_ib_j\langle \vv_i,\vv_j \rangle}{(\sum_{i} a_i)(\sum_{i} b_i )}\geq \frac{\sum_{ij} a_i b_j}{(\sum_{i} a_i)(\sum_{i} b_i )}s_n=s_n
\end{align}

Therefore $s_n$ is monotonically increasing and upper-bounded by $1$ as the cosine of the maximal angle. Due to the monotone convergence theorem, it will converge.
The rest of the proof investigates if the sequence $s_n$ converges to 1 and
if so, under which conditions. \Eqref{eq:s_n_is_monotonic} makes it also clear that we can ignore any $(A_i)_{i \not \in I}$, as the factor before $s_n$ can never be lower than 1. All values are non-negative, $\sum_i a_i > 0$, and $\sum_i b_i > 0$.

\textbf{(3)}
We now introduce a complementary sequence $\varepsilon_n = 1 - s_n$.

Using the result from \eqref{eq:s_n+1}, we write $s_{n+1}$ as:
\begin{align}\label{eq:convergence_speed}
     s_{n+1}=\frac{\sum_{ij} a_ib_j\langle \vv_i,\vv_j \rangle}{\|\sum_{i} a_i \vv_i\|\|\sum_{i} b_i \vv_i\|}=  \frac{\sum_{i\neq j} a_ib_j\langle \vv_i,\vv_j \rangle+\sum_{i} a_ib_i\langle \vv_i,\vv_i \rangle}{\|\sum_{i} a_i \vv_i\|\|\sum_{i} b_i \vv_i\|} = \frac{\sum_{i\neq j} a_ib_j\langle \vv_i,\vv_j \rangle+\langle \va,\vb \rangle}{\|\sum_{i} a_i \vv_i\|\|\sum_{i} b_i \vv_i\|},
\end{align}
since we assumed that $\|\vv_i\|=1$. 
Now, we apply $1=s_n + \varepsilon_n$ and $\langle \vv_i,\vv_j \rangle \geq s_n$
\begin{align}
    s_{n+1}\geq \frac{(\sum_{i\neq j} a_ib_j) s_n+\langle \va,\vb \rangle(s_n +\varepsilon_n)}{\|\sum_{i} a_i \vv_i\|\|\sum_{i} b_i \vv_i\|}\geq \frac{(\sum_{i\neq j} a_ib_j) s_n+\langle \va,\vb \rangle(s_n +\varepsilon_n)}{(\sum_{i} |a_i| \|\vv_i\|)(\sum_{i} |b_i| \|\vv_i\|)} = \frac{(\sum_{i\neq j} a_ib_j) s_n+\langle \va,\vb \rangle(s_n +\varepsilon_n)}{(\sum_{i} a_i)(\sum_{i} b_i)}
\end{align}
Here, we applied the triangle inequality and the fact that the matrix entries $a_i$ and $b_i$ are positive.
\begin{align}
    \label{eq:sn_eps_convergence}
    s_{n+1}\geq \frac{(\sum_{i\neq j} a_ib_j)}{(\sum_{i} a_i)(\sum_{i} b_i)}s_n + \frac{\langle \va,\vb \rangle}{(\sum_{i} a_i)(\sum_{i} b_i)}(s_n + \varepsilon_n) = \frac{(\sum_{ij} a_ib_j)}{(\sum_{i} a_i)(\sum_{i} b_i)}s_n + \frac{\langle \va,\vb \rangle}{(\sum_{i} a_i)(\sum_{i} b_i)} \varepsilon_n
\end{align}
As the factor before $s_n$ is equals one, we have:
\begin{equation}
    \label{eq:sn+1=s_n+eps_n}
    s_{n+1} \ge s_n + \frac{\langle \va,\vb \rangle}{(\sum_{i} a_i)(\sum_{i} b_i)} \varepsilon_n 
    \ge s_n + \lambda \epsilon_n,
\end{equation}
where $\lambda = \frac{\epsDot}{L^2} < \frac{\langle \va,\vb \rangle}{(\sum_{i} a_i)(\sum_{i} b_i)}  $, as $\epsDot$ is the minimal dot product of any two column vectors and $\sum_{i} a_i = ||\va||_1 < L $, as $L$ is the maximum matrix L1-norm of any column vector of $A_n$.

\textbf{(4)}
Suppose, $\lim_{n\to\infty} s_n = 1 - \varepsilon^*$, where 
$\varepsilon^* > 0$. Then, for all $\varepsilon_n$, it must be $ \varepsilon_n > \varepsilon^* $.
Using the result from \eqref{eq:sn+1=s_n+eps_n}, we get:
\begin{align}
    s_{n+1} \geq \varepsilon_n \geq 
    s_n + 
    \lambda
    \varepsilon^*
\end{align}
We would therefore have:
\begin{align}
    \lim_{n\to\infty} s_n \geq \lim_{n\to\infty} n \lambda \varepsilon^* \to \infty.
\end{align}
This is a contradiction. Therefore,  
$\varepsilon^* = 0$ and  $\lim_{n\to\infty} s_n = 1$. The matrix entries of $M_n$ could grow to infinity, therefore $M_\infty$ may not be defined.
However, we normalize the product $\bar{M_n} = M_n/\|M_n\|$, then $\|\bar{M_\infty}\|_1 = 1$ and all the columns of $\bar{M_\infty}$ are the same up to a scalar multiple.
\end{proof}

\paragraph{Update to the Theorem 1 and the Proof:} 
In a previous version
of this manuscript, we did not explicitly require the dot-product $\langle a, b \rangle \ge \epsDot $ to be greater than a constant, 
and also missed a similar constraint for the zero vector.
As a result, 
we did not exclude cases where the matrix sequence $A_n$ converged to such edge-cases. For example, the sequence $A_n = \begin{pmatrix}
    1 & 0  \\
    \frac{1}{n} & 1
\end{pmatrix}
$ fulfilled the criteria of the previous version but is now excluded. 
We also made the normalization more explicit.
An inaccurate statement about the "convergence of the matrix chain $A_n$", which is not actually required, is now removed.
Furthermore, the updated version now contains a
new derivation of the convergence speed.
We thank Günter Rote for bringing these issues to our attention.

\section{Convergence Speed}

For the derivation of the convergence speed, we assume that all  matrices $A_n$ are in the set $I$. We start with the intermediate result from \eqref{eq:sn+1=s_n+eps_n}:
\begin{equation}
    s_{n+1} \geq s_n + \lambda \varepsilon_n = s_n + \lambda (1 - s_n) = (1 - \lambda) s_n + \lambda.
\end{equation}

We now define a new sequence $ \xi_{n+1} = (1-\lambda) \xi_n + \lambda $  with $\xi_0 = s_0$
that is a lower bound of $s_{n+1}$, i.e. for all $s_n \ge \xi_n$

Analyzing the first steps of $\xi_n$, we have:
\begin{align}
\xi_{1} &= (1-\lambda) s_0 + \lambda  \\
\xi_{2} &= (1-\lambda) \left( (1-\lambda) s_0 + \lambda \right) = 
(1-\lambda)^2 s_0 +  (1-\lambda) \lambda  \\
\xi_{3} &= (1-\lambda) \left( (1-\lambda)^2 s_0 +  (1-\lambda) \lambda  \right) = 
(1-\lambda)^3 s_0 +  (1-\lambda)^2 \lambda  
\end{align}
The general form of $\xi_n$ is therefore:
\begin{equation}
\xi_n = (1 - \lambda)^n s_0 + \lambda \sum_{k=0}^{n-1} (1 - \lambda)^k.
\end{equation}
The sum of a geometric series \( \sum_{k=0}^{n-1} r^k \) is given by $\frac{1 - r^n}{1 - r} $.
Applying this to the term of our series $\xi_n$:
\begin{equation}
\lambda \sum_{k=0}^{n-1} (1 - \lambda)^k = \lambda \left( \frac{1 - (1 - \lambda)^n}{1 - (1 - \lambda)} \right) = 1 - (1 - \lambda)^n.
\end{equation}
Therefore, the closed-form solution for $ \xi_n $ is:
\begin{equation}
\xi_n = 1 - (1 - \lambda)^n (1 - s_0).
\end{equation}

Therefore, $s_n$ must also convergence exponentially fast in $\lambda$. 
We bounded $\lambda = \frac{\epsDot}{L^2}$ by the minimum dot-product $\epsDot$ and the maximum L1-norm of the columns of $A_n$. 
In the next section, we conduct an empirical analysis and find that the value of $\lambda$ is large enough that typical matrices converge towards a rank-1 matrix up to floating-point impression within a few steps.

\begin{figure*}[t]
     \centering
    \begin{subfigure}[t]{0.45\textwidth}
        \includegraphics[width=1\textwidth]{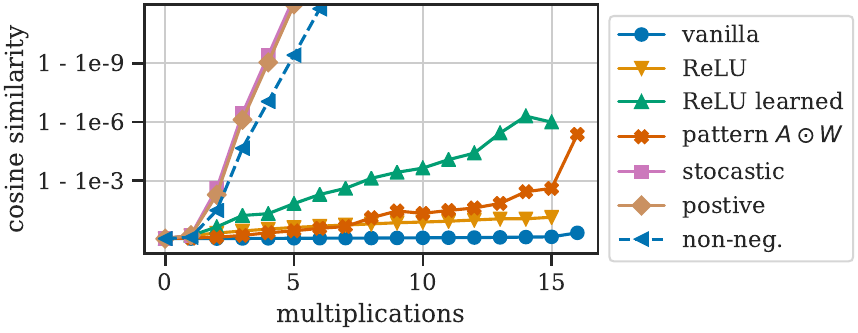}
        \caption{different matrix properties (see text)}
        \label{fig:emp_conv_sq_nn}
    \end{subfigure}
    \hspace{0.8cm}
    \begin{subfigure}[t]{0.45\textwidth}
        \includegraphics[width=1\textwidth]{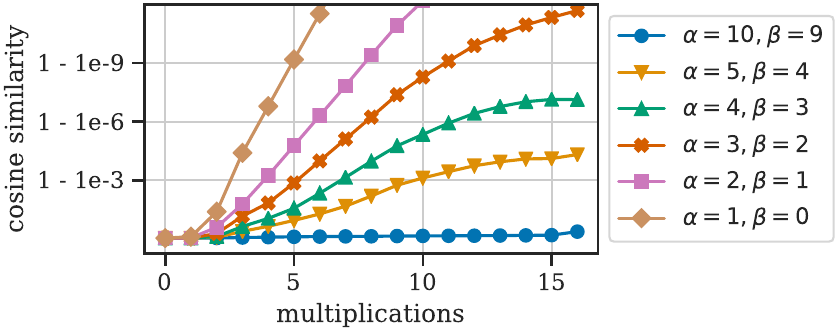}
        \caption{$\alpha W^+ + \beta W^-$ where $W_\idx{ij} \sim \mathcal{N}(0, 1)$}
        \label{fig:emp_conv_ab}
    \end{subfigure}

    \caption{Simulated convergence for a matrix chain.}
    \label{fig:emp_conv}
\end{figure*}

\paragraph{Empirical evaluation} We further investigate the convergence speed using a simulation of random matrices and find that non-negative matrices decay exponentially fast towards 1.

We report the converging behavior for matrix chains which resembles a VGG-16. As in the backward pass, we start from the last layer.
The convolutional kernels are considered to be 1x1, e.g. for a kernel of size (3, 3, 256, 128), we use a matrix of size (256, 128).

We test out the effect of different matrix properties. For \emph{vanilla}, we sample the matrix entries from a normal distribution. Next, we apply a \emph{ReLU} operation after each multiplication.
For \emph{ReLU learned}, we used the corresponding learned VGG parameters.
We generate \emph{non-negative} matrices containing ~ 50\% zeros by clipping random matrices to $[0, \infty]$. And \emph{positive} matrices by taking the absolute value. We report the median cosine similarity between the column vectors of the matrix.

The y-axis of Figure \ref{fig:emp_conv_sq_nn} has a logarithmic scale. We observe that the positive, stochastic, and non-negative matrices
yield a linear path, indicating an exponential decay of the form: $1 - \exp(-\lambda n)$. The 50\% zeros in the non-negative matrices only result in a bit lower convergence slope. After 7 iterations, they converged to a single vector up to floating point imprecision.

We also investigated how a slightly negative matrix influences the convergence. In Figure \ref{fig:emp_conv_ab}, we show the converges of matrices: $\alpha W^+ + \beta W^-$ where $W^+ = \max(0, W), W^- = \min(0, W)$ and $W \sim \mathcal{N}(0, I)$. We find that for small enough $\beta < 4$ values the matrix chains still converge. This simulation motivated us to include \alphaBetaLRP{5}{4} in our evaluation which show less convergence on VGG-16, but its saliency maps also contain more noise.

\section{Pattern Attribution}
\label{appendix:patternattr}

We derive equation \ref{eq:pattern_computed} from the original equation given in \citep{kindermans2018learning}.
We will use the notation from the original paper and denote a weight vector with $\vw = W_{l_\idx{i, :}}$ and the corresponding pattern with $\va = A_{l_\idx{i, :}}$. The output is $y = \vw^T \vx$.

\paragraph{Derivation of Pattern Computation} For the positive patterns of the two-component estimator $S_{\va+-}$, the expectation is taken only over $\{\vx | \vw^T \vx > 0\}$.
We only show it for the positive patterns $\va_+$.
As our derivation is independent of the subset of $\vx$ considered, it would work
analogously for negative patterns or the linear estimator $S_\va$.

The formula to compute the pattern $\va_+$ is given by:
\begin{equation}\label{eq:closed_a+-}
\begin{split}
\va_+ &= \frac{\E_+ \left[\vx y\right]-\E_+ \left[ \vx \right]\E_+ \left[y\right]}
            {\vw^T\E_+ \left[\vx y\right]-\vw^T\E_+ \left[ \vx \right]\E_+ \left[y\right]} \\
      &=\frac{\cov[\vx, \vw^T \vx]}{\vw^T \cov[\vx, \vw^T \vx]},
\end{split}
\end{equation}
where $\cov[\vx, \vw^T\vx ] = \E_+[\vx y] - \E_+[\vx] \E_+[y]$.
Using the bilinearity of the covariance matrix ($\cov[\vb, \vc^T\vd] = \cov[\vb, \vd] \vc$), gives:
\begin{equation}
    \va_+ =\frac{\cov[\vx, \vx]\vw }{\vw^T \cov[\vx,  \vx]\vw}.
\end{equation}
Using the notation $\cov[\vh] = \cov[\vx, \vx]$ gives \eqref{eq:pattern_computed}.

\paragraph{Connection to power iteration} A step of the power iteration is given by:
\begin{equation}
    \vv_{k+1} = \frac{M \vv_k}{\norm{M \vv_k}}
\end{equation}

The denominator in \eqref{eq:pattern_computed} is $ \vw^T \cov[\vh]\vw $.
Using the symmetry of $\cov[\vh]$, we have:
\begin{equation}
    \norm{\cov[\vh]^{1/2} \vw}  = (\vw^T \cov[\vh]^{1/2} \cov[\vh]^{1/2} \vw)^{1/2} = (\vw^T cov[\vh] \vw)^{1/2}
\end{equation}
This should be similar to the norm $\norm{\cov[h]\vw}$.
As only a single step of the power iteration is performed, the scaling should not matter that much.
The purpose of the scaling in the power-iteration algorithm is to keep the vector $v_k$ from exploding or
converging to zero.

\newpage

\section{CIFAR-10 Network Architecture}
\label{appendix:cifar10}
\lstset{language=Python}
\definecolor{codegreen}{rgb}{0,0.6,0}
\definecolor{codegray}{rgb}{0.5,0.5,0.5}
\definecolor{codepurple}{rgb}{0.58,0,0.82}
\definecolor{backcolour}{rgb}{0.95,0.95,0.92}

\lstdefinestyle{mystyle}{
    backgroundcolor=\color{white},
    commentstyle=\color{codegreen},
    keywordstyle=\color{magenta},
    stringstyle=\color{codepurple},
    basicstyle=\ttfamily, 
    breakatwhitespace=false,
    breaklines=true,
    captionpos=b,
    keepspaces=true,
    numbers=none,
    showspaces=false,
    showstringspaces=false,
    showtabs=false,
    tabsize=2
}
\lstset{style=mystyle}

\begin{lstlisting}
    # network architecture as a keras model
    model = Sequential()

    model.add(InputLayer(input_shape=(32, 32, 3), name='input'))
    model.add(Conv2D(32, (3, 3), padding='same', name='conv1'))
    model.add(Activation('relu', name='relu1'))
    model.add(Conv2D(64, (3, 3), padding='same', name='conv2'))
    model.add(Activation('relu', name='relu2'))
    model.add(MaxPooling2D(pool_size=(2, 2), name='pool2'))

    model.add(Conv2D(128, (3, 3), padding='same', name='conv3'))
    model.add(Activation('relu', name='relu3'))
    model.add(Conv2D(128, (3, 3), padding='same', name='conv4'))
    model.add(Activation('relu', name='relu4'))
    model.add(MaxPooling2D(pool_size=(2, 2), name='pool4'))

    model.add(Flatten(name='flatten'))
    model.add(Dropout(0.5, name='dropout5'))
    model.add(Dense(1024, name='fc5'))
    model.add(Activation('relu', name='relu5'))
    model.add(Dropout(0.5, name='dropout6'))
    model.add(Dense(10, name='fc6'))
    model.add(Activation('softmax', name='softmax'))
\end{lstlisting}

\section{Results on ResNet-50}
\label{appendix:result_resnet}

\begin{figure}[H]
    \centering
     \begin{subfigure}[t]{0.35\textwidth}
        \includegraphics[height=5.cm]{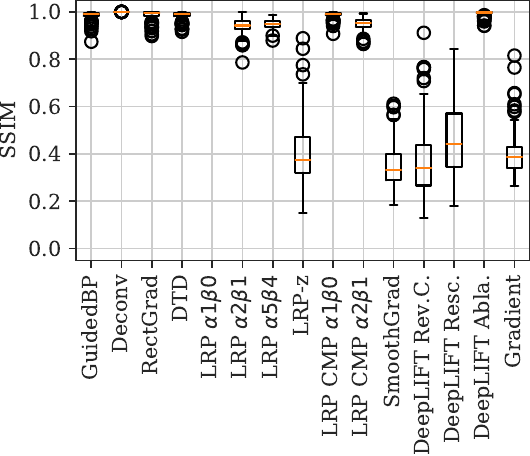}
        \caption{Random Logits}
        \label{fig:random_logits_resnet50}
     \end{subfigure}
     \hspace{0.5cm}
     \begin{subfigure}[t]{0.60\textwidth}
        \includegraphics[height=5.cm]{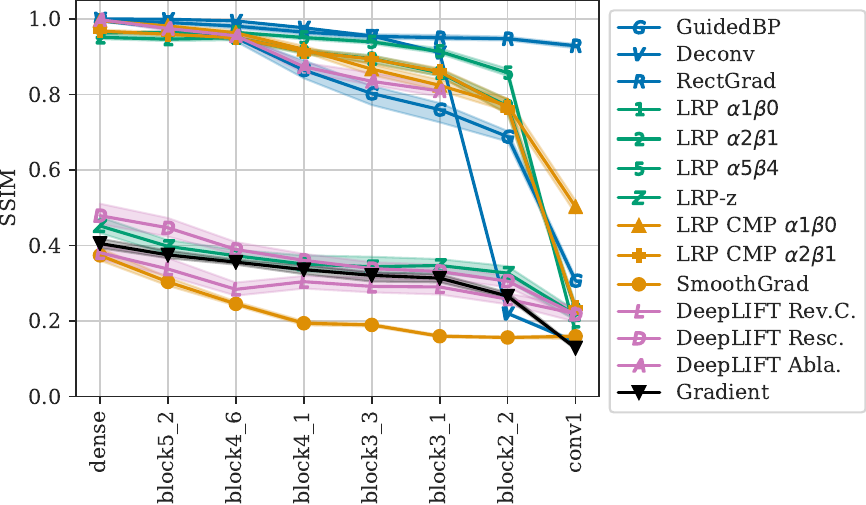}
        \caption{Cascading Parameter Randomization}
        \label{fig:random_params_resnet50}
     \end{subfigure}
     \caption{Effect of (a) randomizing the logits or (b) the parameters on a ResNet-50.}
     \label{fig:sanity_checks_resnet50}
\end{figure}

\newpage

\section{Additional Cosine Similarity Figures}
\label{appendix:csc_figures}

\renewcommand{\cossimHeight}{3.20cm}
\begin{figure}[H]
    \begin{subfigure}[t]{\textwidth}
    \includegraphics[height=\cossimHeight]{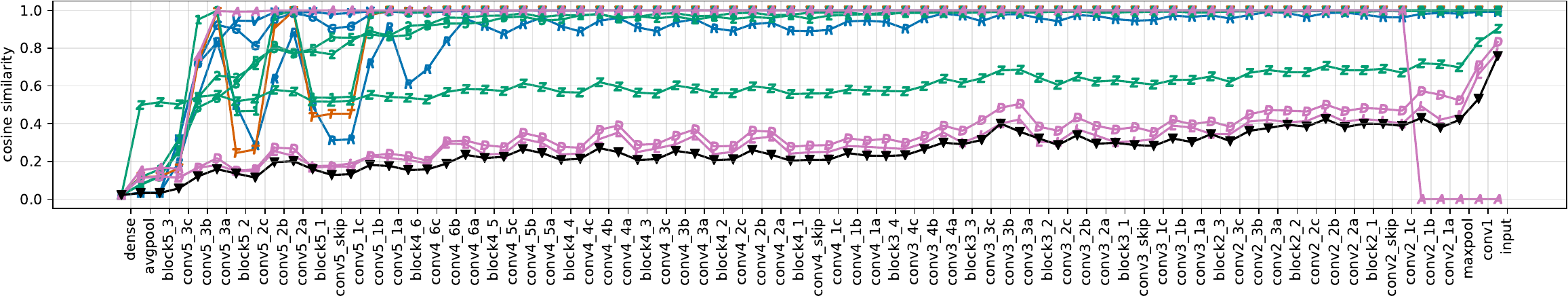}
        \caption{ResNet-50 (linear)}
        \label{fig:csc_resnet50_layer176_linear}
    \end{subfigure}
    \begin{subfigure}[t]{0.6\textwidth}
    \includegraphics[height=\cossimHeight]{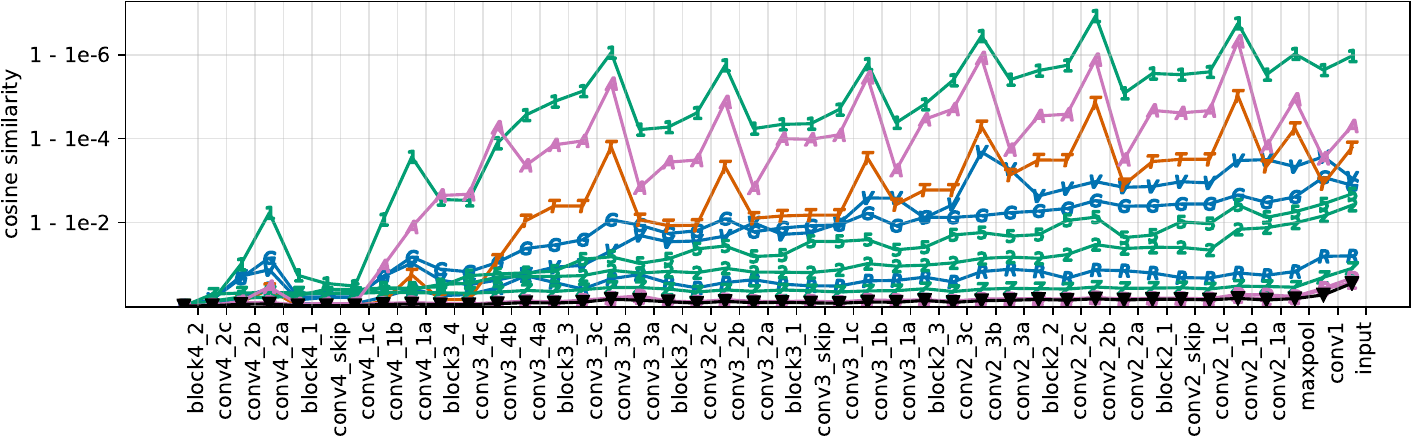}
        \caption{ResNet-50 (log)}
    \end{subfigure}
    \begin{subfigure}[t]{0.4\textwidth}
        \includegraphics[height=\cossimHeight]{figures/cosine_similarity/vgg16/vgg16_layer_22_log.pdf}
        \caption{VGG-16}
    \end{subfigure}
    \begin{subfigure}[t]{0.5\textwidth}
        \includegraphics[height=\cossimHeight]{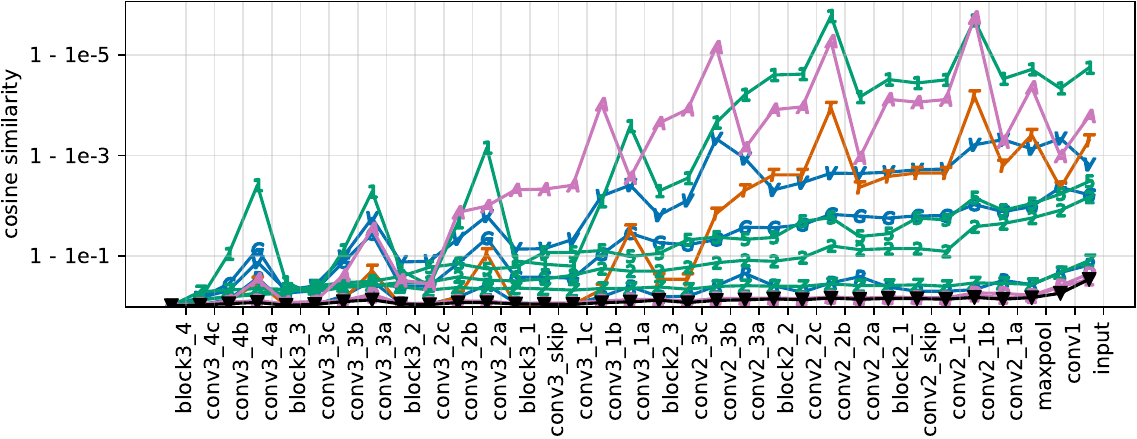}
        \caption{ResNet-50}
    \end{subfigure}
    \begin{subfigure}[t]{0.27\textwidth}
        \includegraphics[height=\cossimHeight]{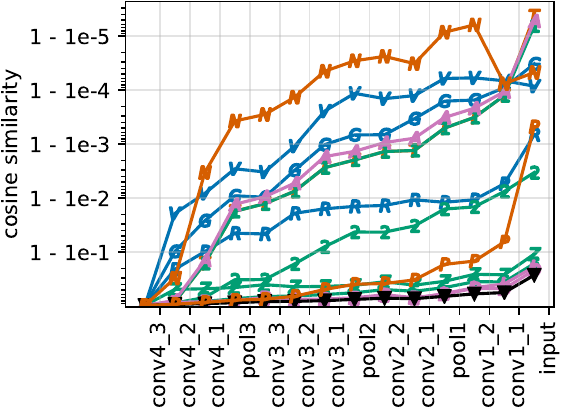}
        \caption{VGG-16}
    \end{subfigure}
    \begin{subfigure}[t]{0.18\textwidth}
        \raisebox{0.0cm}{
            \includegraphics[height=\cossimHeight]{figures/cosine_similarity/cos_sim_legend_all.pdf}
        }
    \end{subfigure}

    \begin{subfigure}[t]{0.25\textwidth}
        \includegraphics[height=\cossimHeight]{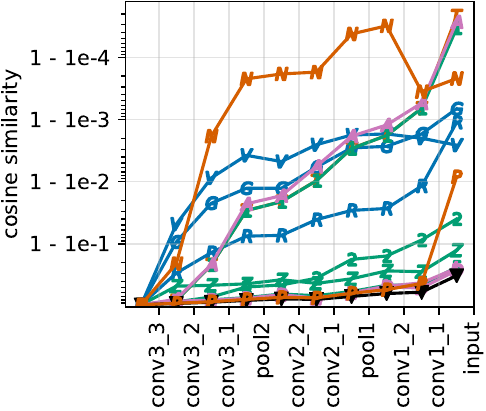}
        \caption{VGG-16}
    \end{subfigure}
    \begin{subfigure}[t]{0.25\textwidth}
        \includegraphics[height=\cossimHeight]{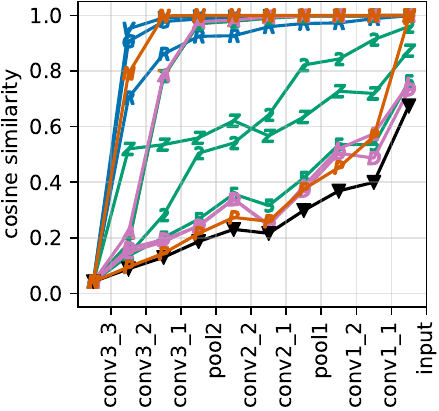}
        \caption{VGG-16 (linear)}
    \end{subfigure}
        \begin{subfigure}[t]{0.25\textwidth}
        \includegraphics[height=\cossimHeight]{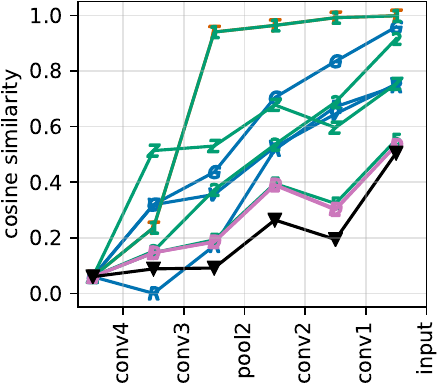}
        \caption{CIFAR-10 (linear)}
    \end{subfigure}

    \caption{Convergence measured using the CSC for different starting layers.}
    \label{fig:appendix_cos_sim}
\end{figure}

\newpage

\section{Saliency maps for Sanity Checks}
\label{appendix:sanity_checks}

For visualization, we normalized the saliency maps to be in $[0, 1]$ if the method produce only positive relevance.
If the method also estimates negative relevance, than it is normalized to $[-1, 1]$. The negative and positive values are scaled equally by the absolute maximum.
For the sanity checks, we scale all saliency maps to be in $[0, 1]$.

\begin{figure}[H]
    \centering
    \begin{subfigure}[t]{0.49\textwidth}
        \includegraphics[width=\textwidth]{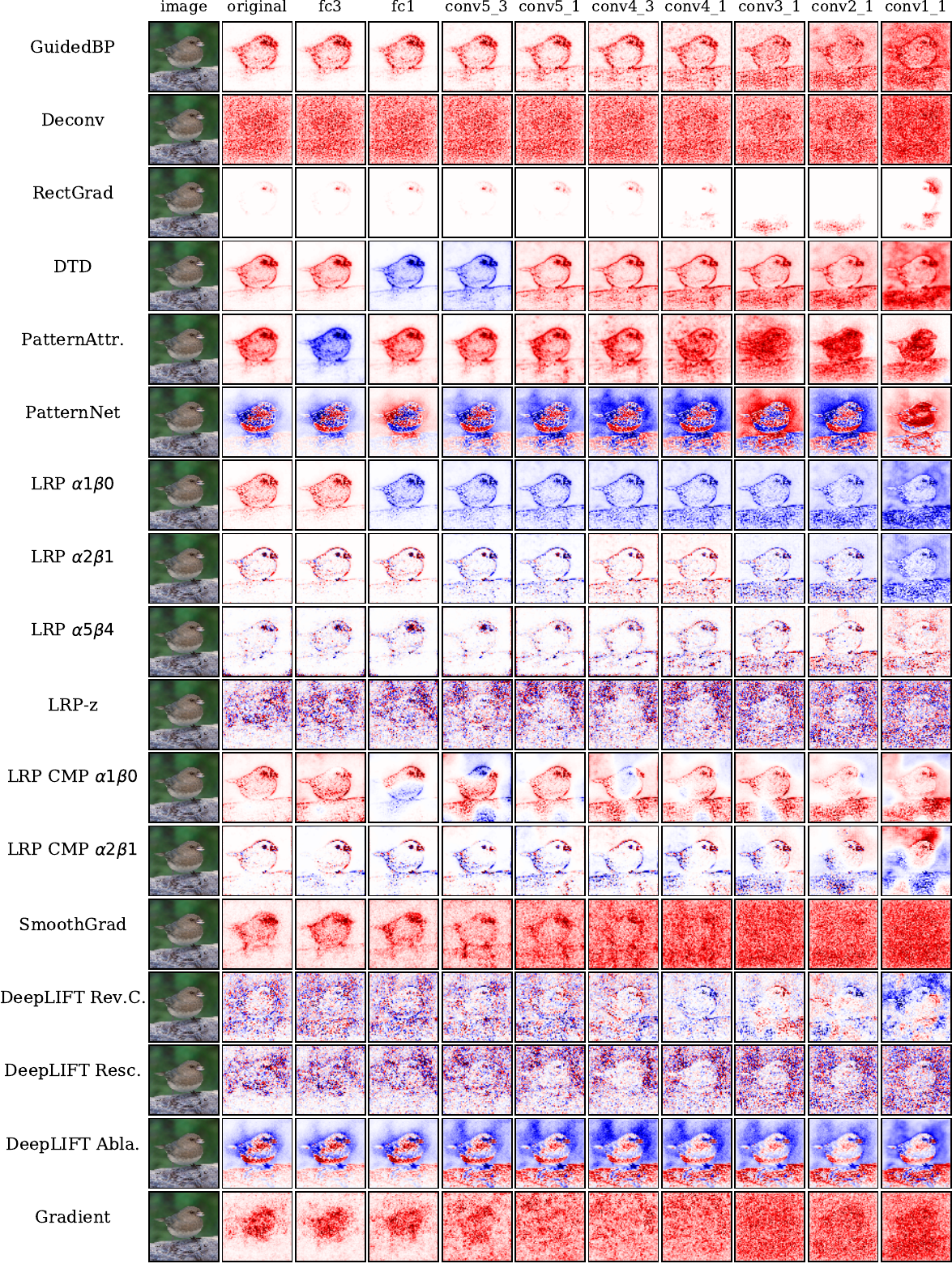}
        \caption{VGG-16. }\label{fig:sanity_checks_vgg16}
    \end{subfigure}
    \begin{subfigure}[t]{0.49\textwidth}
         \includegraphics[width=\textwidth]{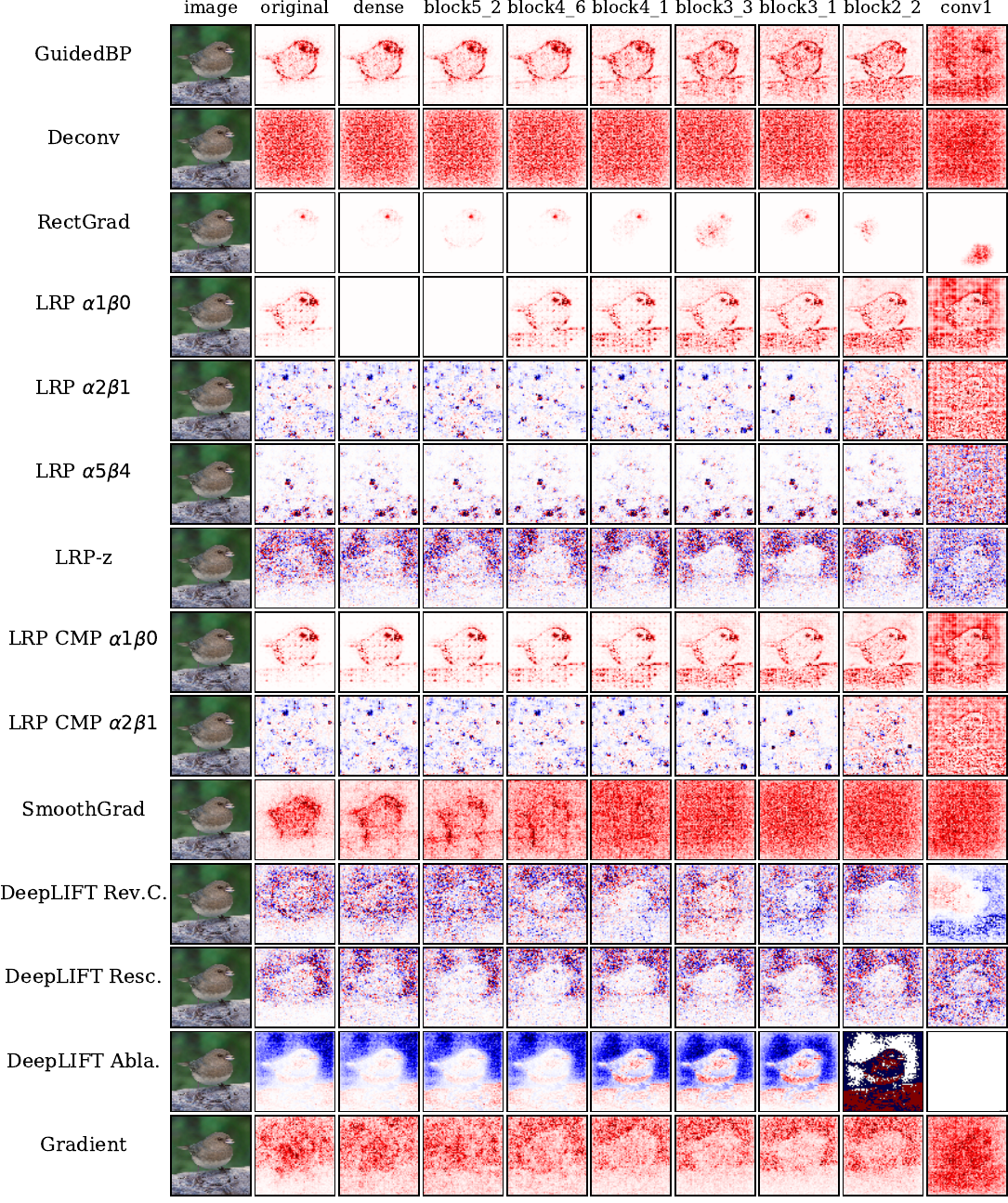}
        \caption{ResNet-50. }
        \label{fig:saliency_resnet}
    \end{subfigure}
    \caption{Saliency maps for sanity checks. Parameters are randomized starting from last to first layer.}
\end{figure}

\end{document}